\newif\ifabstract
\abstracttrue
\abstractfalse 
\newif\iffull
\ifabstract \fullfalse \else \fulltrue \fi

\documentclass[11pt]{article}
\usepackage{amsfonts}
\usepackage{amssymb}
\usepackage{amstext}
\usepackage{amsmath}
\usepackage{amsthm}
\usepackage{algorithm}
\usepackage{algorithmic}
\usepackage{tikz}
\usepackage{multicol}
\usetikzlibrary{calc}
\usetikzlibrary{arrows,shapes}
\usepackage{dot2texi}
\usepackage{siunitx}
\usepackage{booktabs}
\usepackage{pdflscape}
\usepackage{threeparttable}
\usepackage{longtable}
\usepackage{multirow}
\usepackage{blindtext}
\usepackage[inline]{enumitem}
\usepackage{xcolor}

\usepackage{xspace}
\usepackage{theorem}
\usepackage{graphicx}
\usepackage{url}
\usepackage{graphics}
\usepackage{colordvi}
\usepackage{colordvi}
\usepackage{lscape}
\usepackage{natbib}
\usepackage[english]{babel}
\usepackage{float}
\usepackage{caption}
\usepackage{subcaption}
\usepackage{geometry}
\usepackage{amsmath}

\advance \oddsidemargin by -0.8in
\newcommand{\myparskip}{3pt}
\parskip \myparskip

{\hspace*{\fill}$\Box$\par\vspace{4mm}}

\begin{document}

\title{An investigation into machine learning approaches for forecasting spatio-temporal demand in ride-hailing service \footnote{Currently under review for publication}}
\author{Isma\"il Saadi \thanks{Local Environment Management \& Analysis (LEMA), Department of Urban and Environmental Engineering (UEE), University of Li\`ege, All\'ee de la D\'ecouverte 9, Quartier Polytech 1, Li\`ege, Belgium, Email: {ismail.saadi@ulg.ac.be}} \and Melvin Wong \thanks{Laboratory of Innovations in Transportation (LITrans), Department of Civil, Geotechnical, and Mining Engineering, Polytechnique Montr\'eal, Montr\'eal, Canada, Email: {melvin.wong@polymtl.ca}} \and Bilal Farooq \thanks{Laboratory of Innovations in Transportation (LITrans), Department of Civil, Geotechnical, and Mining Engineering, Polytechnique Montr\'eal, Montr\'eal, Canada, Email: {bilal.farooq@polymtl.ca}} \and Jacques Teller \thanks{Local Environment Management \& Analysis (LEMA), Department of Urban and Environmental Engineering (UEE), University of Li\`ege, All\'ee de la D\'ecouverte 9, Quartier Polytech 1, Li\`ege, Belgium} \and Mario Cools \thanks{Local Environment Management \& Analysis (LEMA), Department of Urban and Environmental Engineering (UEE), University of Li\`ege, All\'ee de la D\'ecouverte 9, Quartier Polytech 1, Li\`eège, Belgium}}

\begin{titlepage}
\maketitle

\thispagestyle{empty}

\begin{abstract}
In this paper, we present machine learning approaches for characterizing and forecasting the short-term demand for on-demand ride-hailing services. We propose the spatio-temporal estimation of the demand that is a function of variable effects related to traffic, pricing and weather conditions. With respect to the methodology, a single decision tree, bootstrap-aggregated (bagged) decision trees, random forest, boosted decision trees, and artificial neural network for regression have been adapted and systematically compared using various statistics, e.g. R-square, Root Mean Square Error (RMSE), and slope. To better assess the quality of the models, they have been tested on a real case study using the data of DiDi Chuxing, the main on-demand ride-hailing service provider in China. In the current study, 199,584 time-slots describing the spatio-temporal ride-hailing demand has been extracted with an aggregated-time interval of 10 mins. All the methods are trained and validated on the basis of two independent samples from this dataset. The results revealed that boosted decision trees provide the best prediction accuracy (RMSE=16.41), while avoiding the risk of over-fitting, followed by artificial neural network (20.09), random forest (23.50), bagged decision trees (24.29) and single decision tree (33.55).
\end{abstract}

\end{titlepage}

\section{Introduction}
\label{S:1}

Shared mobility is increasingly becoming a dominant mode of transportation with substantial interest from urban population as well as transportation agencies. Typically, two or more travelers may share a vehicle for a trip in order to reduce the travel costs, inconvenience of driving, and to have the ability to do tasks other than driving (e.g. working, using smartphone, etc.). On-demand ride-sharing companies generally provide services which are more convenient and flexible than traditional public transportation systems \citep{Furuhata201328}.

With respect to the environmental aspects, ride-sharing can have a positive effect on the GHG emission reductions, as the total number of trips in urban areas may reduce. In addition to reduced travel costs, \citet{Morency2007} stated that maximizing car occupancy can result in a reduction of the number of vehicles traveling on urban roads, as well as traffic congestion. Thanks to shared mobility, significant improvements can be carried out when it comes to dealing with the mitigation of wasted travel time and fuel consumption. In fact, ride-sharing can be perceived as a system of co-utility \citep{Sanchez2016147} as all the parties involved within the sharing process have some potential benefit at stake.

Historically, the concept of shared mobility has started with the emergence of carpooling around the 1970’s. Generally, group of people would agree on sharing the same vehicle for performing similar trips in order to reduce travel costs, fuel consumption and tolls. Also, the stress of driving is mitigated and more parking spaces are available.

Nowadays, following the development in terms of Information and Communication Technologies (ICT)  and the ubiquitousness of smartphones, on-demand ride-hailing service providers tapped into this high-potential market in order to better structure the shared-mobility systems for profit maximization. Different Internet and mobile app-based platforms, i.e. Uber, Lyft, and DiDi Chuxing, propose a wide range of dynamic ride-hailing services. They consist of matching drivers and travelers according to very short time intervals or even en-route. Due this flexibility and range of services, the number of registered users in ride-hailing platforms, e.g. Kuaidi (a competitor of DiDi) or DiDi, have reached 60 and 100 million in China \citep{he2015modeling}. Taking into account the further proliferation of smartphone-based applications in near future \citep{he2015modeling}, this will probably result in a significant increase in the demand for ride-hailing. Therefore, an important part of the ride-hailing literature is dedicated to the development of matching algorithms where the objective is to reach the most optimal match between drivers and the requests of the travelers depending on geographical locations, travel costs and other additional factors \citep{Agatz2012295}. Typically, on-demand requests from travelers are launched through mobile apps. The users have the possibility of selecting the most appropriate shared-trip option depending on price and time. Nearby drivers are prompted of this request and one of them can accept the request for ride-hailing.

Although huge efforts have been made by researchers for optimizing the matching between drivers and travelers on a real time basis using highly sophisticated algorithms, these are generally focusing on operational aspects within very specific scope. The observations stemming from on-demand ride-hailing service providers revealed that in case of rush/peak-hours, the requests coming from the travelers are more difficult to fulfill compared to the periods with low-traffic rate, during off-peak hours or within districts with low ride-hailing demand levels.

In this context, the objective of the current study is to quantitatively determine the short-term spatial, i.e. at district level, and temporal, i.e. day of week and time of day, distribution of the ride-hailing demand as a function of various variables related to weather, traffic conditions and pricing. To achieve this goal, we first classify the set of available variables and select the most relevant ones based on a feature selection technique, RreliefF \citep{Kononenko1997}. Regression based machine learning techniques are then presented and calibrated for accurately forecasting the short-term demand for ride-hailing. The different approaches are also compared on the basis of key statistics to identify which one provides the best generalized predictive capabilities for the case of short-term demand prediction. Ride-hailing demand estimation is a key piece of information for on-demand service providers. By anticipating the demand within short-term intervals, the ride-hailing service providers can better fulfill the requests of the users. In doing so, the imbalance between the demand and supply can be mitigated preemptively.

Rest of the article is organized as follows: first we present the existing state of literature associated to the topic. We present the data and descriptive analysis. Modeling technique are described, based on which results are developed and detailed comparative analysis is performed. In the end we discuss the conclusions and future direction.
\section{Literature review}
\label{S:2}

Based on the destination and route, shared mobility services can broadly be classified into one of the four types as shown in Table \ref{table types of shared}. Traditional public transport falls into the fixed route and destination category where passengers are picked up at designated stations along a fixed route. Floating car-sharing services have a non-fixed route and non-fixed destination system where users can return the vehicle to any open lots (within a serviceable zone) and to take any route they desire. Taxis, and dynamic ride-hailing options such as Uber, DiDi or Lyft have a fixed pre-stated destination (given by the passenger) but a non-fixed route, i.e. drivers can choose whichever route to take. Bicycle- and car-sharing systems operate on a centralized hub where users rent a vehicle for a limited time period before returning to the same or different hub. Recent autonomous transport services (e.g. NuTonomy in Singapore) falls into the non-fixed destination, but fixed routes category \citep{anselmetti2016}.

\begin{table}[H]
	\centering
	\begin{tabular}{|r|r|r|}
		\hline
		& Fixed route & Non-fixed route \\
		\hline
		\multirow{2}{*}{Fixed destination} & buses & taxis \\
		& metro/subway & ride-hailing\\
		\hline
		\multirow{2}{*}{Variable destination} & \multirow{2}{*}{Autonomous taxis} & car-sharing\\
		&  & bike-sharing\\
		\hline
	\end{tabular}
	\centering\caption{Classification of the different types of shared mobility services.}
	\label{table types of shared}
\end{table}

In order to pro-actively fulfill the requests from travelers, on-demand ride-sharing service companies need to forecast the spatio-temporal dynamics of the demand. In doing so, the on-demand service provider can operate the supply side more efficiently. For instance, with respect to the daily commuters, ride-hailing requests are particularly important in the mornings in residential areas and in the evenings at work places/business centers. Thus, more cars should be available in specific locations at a specific period of time while reducing the supply for other places.

Unfortunately, the existing literature on demand estimation in ride-hailing systems is extremely rare in transportation research. Most of the research studies are focusing on operational aspects of ride-sharing including the matching between drivers and riders and their related optimization aspects. Also, other studies try to incorporate policy effects within the models to determine what are the most adapted choices in terms of fleet sizing, ride-sharing options or fleet locations.

\citet{ciari2013} proposed a policy-sensitive activity-based model including agent-based aspects for car-sharing demand estimation. Simple rules are defined at micro-scale for the agent's behavior. The interactions between the synthetic agents lead to a stable macro-behavior of the virtual system. They managed to reproduce some typical usage patterns of ride-sharing. Similarly, \citet{balac2015carsharing} proposed a car-sharing demand estimation model based on the agent-based modeling framework MATSim \citep{horni2016multi}. Effects of the supply changes on the demand have been investigated on both spatial and temporal basis. However, their study is mainly focusing on the selection of the most optimal alternatives, i.e. one-way station or round trip, taking into account the fleet size. Unfortunately, it does not really provide a methodology capable of predicting short-term demand for car-sharing for specific period of time and space.

A significant number of short-term prediction traffic demand forecasting models have been developed in theoretical and applied intelligent transportation systems (ITS). These models are generally used to simulate the actual traffic characteristics and also predict the future traffic conditions. Overall, the models consider travel volume, speed, density characteristics and travel times as predicted outcomes. Note that in our study, the outcome is the future number of requests, i.e. the demand, according to space and time and other additional external factors. In this paper, we are rather seeking to provide a brief overview of the techniques that have been used in short-term traffic demand forecasting with an emphasis on the methodological aspects. Indeed, a large number of studies related to the technological or analytical aspects of short-term traffic demand prediction can be found in the literature. For instance, one can refer to the review paper of \citet{vlahogianni2014short} in order to get a global overview of the state of the art.

In the early days of the short-term traffic demand prediction, most of the studies were mainly using "conventional" statistical-based approaches to predict the anticipated state of the traffic at a single point. Recently, with the emergence of big data and developments in ICT, algorithms and computational processes, the research has shifted toward artificial intelligence (AI)-based approaches including Bayesian Networks (BN), Artificial Neural Networks (ANN), Fuzzy Logic (FL) or Evolutionary-based Algorithms (EA) \citep{vlahogianni2014short}. Studies have proved that the latter techniques provide more accurate predictions than the "conventional" statistical-based approaches and especially in the context of unstable traffic conditions, complex road network settings and the size and complexity of the actual datasets.

In terms of ride-hailing demand estimation problem, to our knowledge, no studies have succeeded in incorporating spatio-temporal dynamics. In a very recent study, \citet{li2017taxi} developed a wave-SVM model to predict short-term traffic demand as a data-driven approach. The advantages of wavelet and Support Vector Machine (SVM) have been combined in one framework to mitigate the negative effects induced by both approaches. Their methodology is based on four successive steps: the original dataset, stemming from the on-demand ride-hailing service provider DiDi, is pre-processed to describe the time series of ride-hailing demand. Then, the time series is decomposed into independent low-high frequency time series using the wavelet technique. Based on the frequency-based clustering of the time series, a set of SVM-based models is trained for each cluster to predict their respective time series. Finally, by reconstructing the predicted time series using once again the wavelet technique, the simulated demand is obtained. The model has been tested on a real case study in China using the same dataset of the current article. The validation is only performed on a single day and results show that their approach was over-fitted for morning peak-hour, while the errors were high for rest of the day. \citet{li2017taxi} have highlighted that the wave-SVM model is capable of capturing the non-stationary characteristics although the system presents many instabilities and non-linearities, especially for short-term predictions.

Wave-SVM model presents some important shortcomings. (a) The predictions do not incorporate specific space-time patterns. Indeed, the proposed approach is only capable of predicting the ride-hailing demand based on time without differentiating weekdays and week-ends which clearly have different behavioral patterns (see Figure \ref{F:2}). In addition, (b) the transition from a district to another also presents some important differences with respect to the number of requests. Generally, for districts with low number of requests, the demand patterns are difficult to capture and they include more spread. In contrast, for districts with high number of requests, the demand patterns are more visible and easier to capture for time series-based techniques. In this context, a wave-SVM strategy is only valid for a limited number of districts and cannot be generalized for the entire city. Also, (c) the wave-SVM model is not sensitive to any external factors such as traffic levels within districts, weather conditions or pricing.

Models for short-term traffic demand prediction can be classified into three separate categories \citep{li2017taxi}. First, the linear models including the "conventional" statistical-based approaches discussed above and ARIMA for time series predictions. Then, the non-linear models which provide more accurate results since non-linear effects in transportation systems can be taken into account. The third category includes the machine learning or data-driven methods which have been received with a lot of interest in short-term traffic forecasting \citep{li2017taxi,vlahogianni2014short}. Finally, in case of complex traffic forecast configurations, combinations of the above mentioned technique can be adopted. 

In the current study, we propose, as a main contribution, an integrated modeling framework based on regression based nonparametric machine learning techniques for forecasting the short-term ride-hailing demand according to space, i.e. at district level, and time, i.e. day of week and time of day. Contrary to the model of \citet{li2017taxi}, the variable effects with respect to weather, traffic conditions, current demand level, and pricing are also included into the current integrated model. In doing so, higher levels of predictive capability can be achieved. We also show how the proposed integrated framework better reproduces the irregular demand patterns for low-demand level districts over different days of the week.

In particular, we propose to forecast the short-term ride-hailing demand by using four regression based machine learning methods: (a) single decision tree, (b) bagged decision trees, (c) Random Forest and (d) boosted decision trees. The short-term demand forecasting for ride-hailing is considered as a pattern recognition problem contrary to the model of \citet{li2017taxi} who considered it from a time series perspective. As highlighted previously, these data-driven-based approaches are more adapted for identifying and extracting the demand patterns from large and complex datasets. In addition, this category of models can handle categorical predictors which are characterized by an important number of levels. 
Compared to the discrete choice models, machine learning-based techniques are more flexible and have more degrees of freedom, which are necessary for short-term prediction. Thus, more complex distribution shapes can be fitted. The best decision tree-based algorithm will be compared with a Artificial Neural Network (ANN) in order to provide deeper insights in terms of performances and predictive capabilities.

As machine learning techniques are adopted in the current study, this choice raises some key issues that should clearly be highlighted and discussed. For example, how should one assess the quality of a model and to what extent should the error properties of the same model be? Although a model may demonstrate a good fit with respect to the observed dataset, it does not mean that the prediction accuracy will also be good. Sometimes, some irregular patterns are not present in the training dataset or the model may be over-fitted, then, the model fails to reproduce or characterize such irregular patterns in other circumstances. The latter aspects will be addressed rigorously by examining the level of generalization of each model.

\section{Data}
\label{S:3}

DiDi Chuxing is considered as the world’s largest on-demand ride-hailing service platform \citep{chen2017understanding}. DiDi proposes a wide range of services including taxi-hailing in 360 cities, private car-hailing in 80 cities, social riding (Hitch) in more than 300 cities, DiDi Chauffeur, DiDi Bus, DiDi Test Drive, DiDi Car rental and DiDi entreprise solutions \citep{didi2017}. All the services are proposed through an Internet-based smartphone app. 

The mobility related statistics for ride-sharing in China are quite impressive. Indeed, the services provided by DiDi have reached around 300 million users across China with 3 million rides per day for taxi-hailing only. The peer-to-peer private car services have reached 3 million rides as well, which also makes it as one of the most successful ride-sharing option after taxi-hailing. With respect to the Hitch services, DiDi revealed that it managed to reach around 600,000 rides per day since its introduction within a period of one month. This further shows the promising potential of sharing mobility services in that specific market. Also, some additional services, for instance designated driving and bus services, have been proposed recently. Besides, in order to meet the constant mobility needs in China and to solve the underlying problems associated to congestion and pollution, DiDi is always improving the quality, the flexibility and the efficiency of their services. For example, within the three coming years, DiDi aims at serving around 30 million of users and 10 million drivers on a daily-basis. Another objective of DiDi is to propose a ride-sharing system that is capable of picking up a passenger from any geographical location at any time within a delay of 3 minutes \citep{didi2017}. The predictive models developed in this article could be very useful in achieving such objectives.


In the current study, we use the transactions data provided by the on-demand ride-hailing service provider DiDi in an undisclosed city of China, covering a period of three weeks from the 1st to the 21st of January 2016. The individual requests, i.e. demand, are registered across 66 anonymized districts, seven days of the week. Based on DiDi's initial request, we chose a prediction interval of 10 minutes. So, the observed requests are aggregated within a time interval of 10 minutes from 00:00 am to 11:59pm, each day. Resulting dataset consists of 199,584 observations over times-lots equally distributed across each recorded weeks (66,528 observations/week) and days (9,504 observations/day). Similarly, the observations were distributed across 66 districts with 3,024 observations/district. In addition to the 10-minutes aggregated demand, the dataset also provides the fulfilled demand which can also be qualified as the "supply". It can be processed to have the same spatial and temporal resolution.

The basic statistics, i.e. average, maximum, minimum and median, related to pricing are also provided within each time-slot. Traffic conditions are represented in form of four Levels of Service (LoS), where 1 is least congested and 4 is most. This LOS is defined at local scale (district) and global scale (city). DiDi provided the percentage of vehicles experiencing each LoS in a district and also at city level for each time-step. Finally, three important indicators related to weather conditions are also included, but with an aggregated temporal resolution of 180 minutes. The weather conditions are represented by a weather index, the temperature and the PM2.5 level (an air quality index).

\begin{figure}[H]
	\centering
	\includegraphics[width=\textwidth]{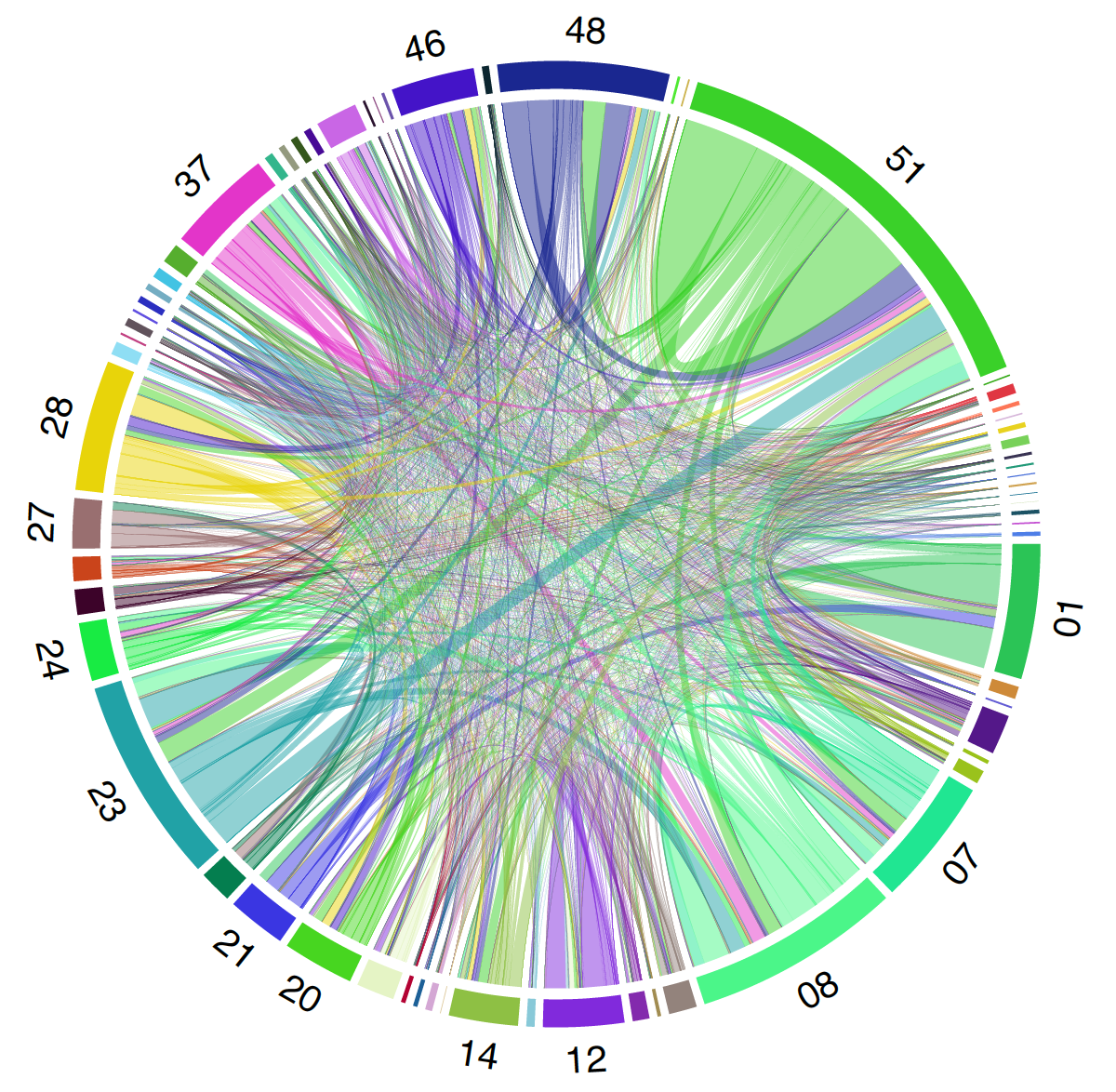}
	\caption{Distribution of the O-D flows of the ride-hailing system}
	\label{fig:001}
\end{figure}

Figure \ref{fig:001} presents the O-D flows between all the anonymized districts of the city. The importance of the flows is translated by the width of the link. We have only labeled the major districts which generate and attract highest number of riders. One could depict from the chord diagram that district 51 presents the most important ride-hailing activity. Also, important ride-hailing trips are taking place within the same district. A smaller flow goes towards district 24. In addition, the ride-hailing activity between the districts is quite heterogeneous. A large number of districts with small demand patterns can be observed.

\begin{figure}[H]
	\centering
	\begin{subfigure}[b]{0.45\textwidth}
		\includegraphics[width=\textwidth]{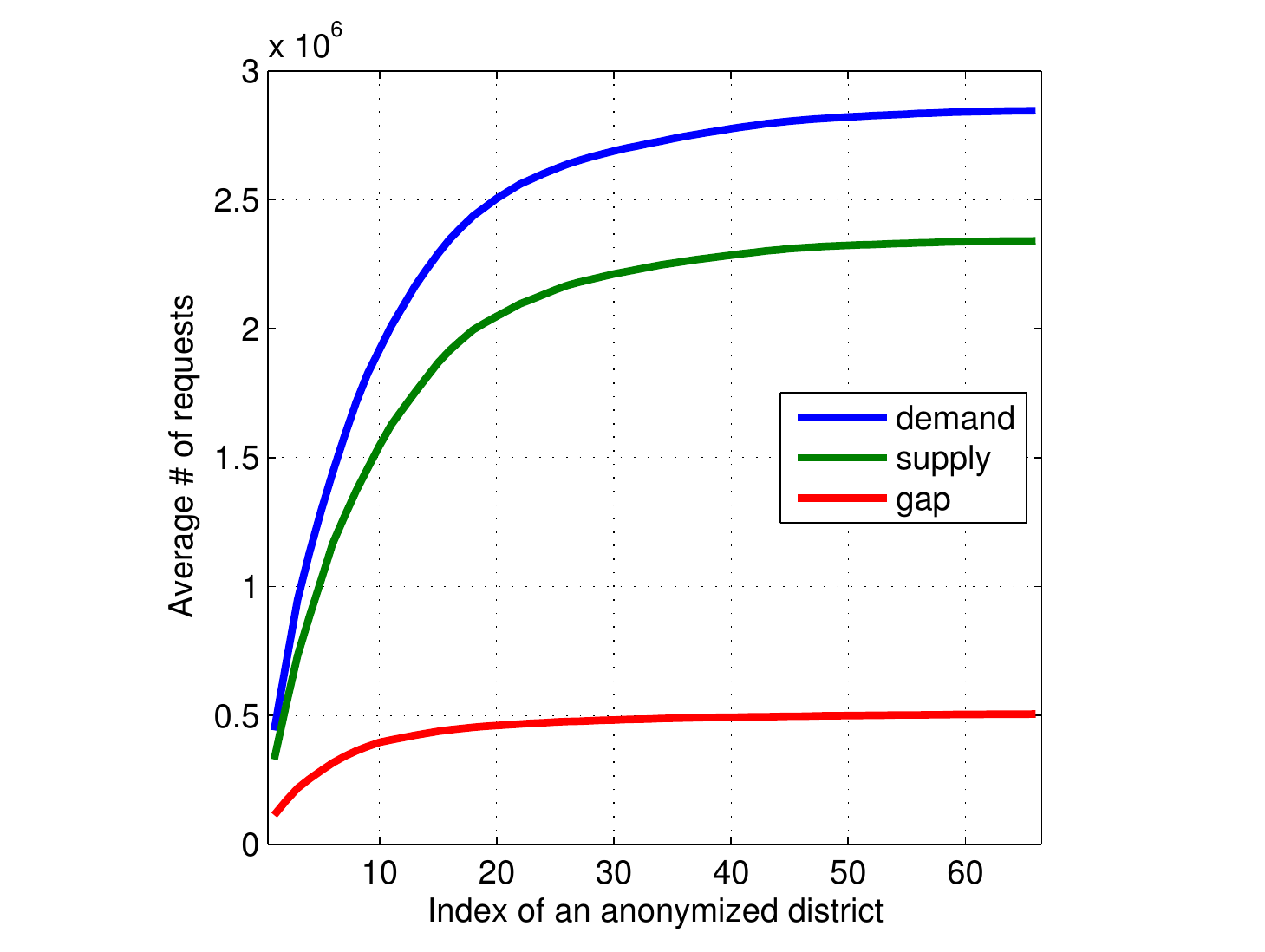}
		\caption{Cumulative distribution of the demand, the supply and the gap across the districts}
		\label{fig:1a}
	\end{subfigure}
	\begin{subfigure}[b]{0.45\textwidth}
		\includegraphics[width=\textwidth]{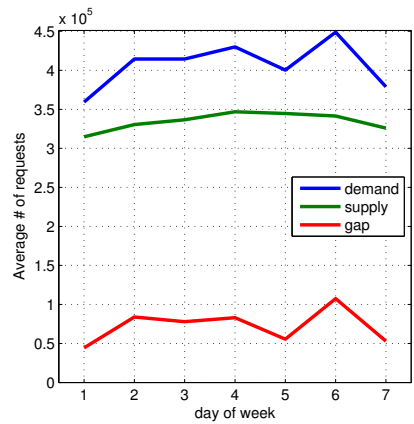}
		\caption{Evolution of the supply, the demand and the gap based on the day of week}
		\label{fig:1b}
	\end{subfigure}
	\begin{subfigure}[b]{0.45\textwidth}
		\includegraphics[width=\textwidth]{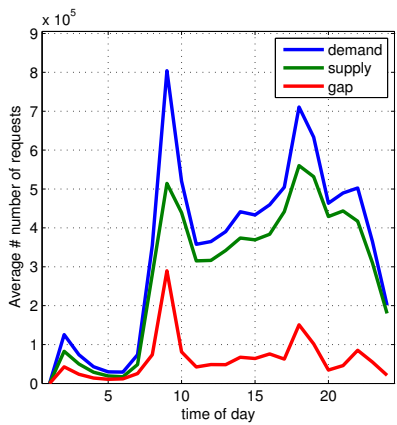}
		\caption{Evolution of the demand-supply during a day}
		\label{fig:3a}
	\end{subfigure}
	\begin{subfigure}[b]{0.45\textwidth}
		\includegraphics[width=\textwidth]{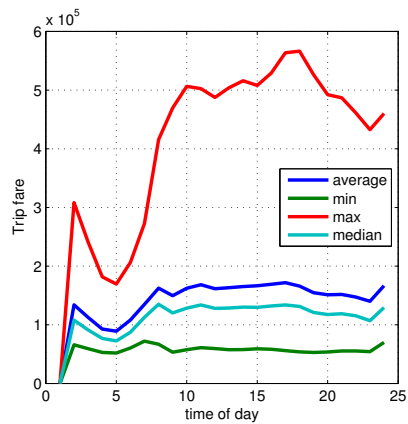}
		\caption{Evolution of the ride costs during the day}
		\label{fig:3b}
	\end{subfigure}
	\caption[]{Demand patterns}
	\label{fig:3}
\end{figure}

Figure \ref{fig:1a} presents the comparison between the cumulative distributions of the demand, the supply and the gap, i.e. difference between demand and supply, over all the districts. It is interesting to observe that the average cumulated number of non-fulfilled requests, i.e. the gap, is around 0.5$\times$10e6. It represents 17.5\% of the average total demand per week over all the districts. In such cases, additional efforts are needed to be carried out by DiDi to fulfill the unmet demand. This outlines the need to develop tools that can better characterize and forecast the short-term demand. Using predictive models, the supply can be dynamically regulated and adapted according to the spatio-temporal characteristics of the ride-hailing demand.

Figure \ref{fig:1b} presents the evolution of the average demand, supply and gap across a full week. One can conclude from the graph that the demand presents variable patterns compared to the supply. For example, on Saturday, the average number of requests is relatively high. Generally, on Saturday people prefer to perform social activities, using ride-hailing as mean of transportation. In contrast, the demand is relatively low on Sunday when a considerable part of people prefer to stay at home. Regarding the supply, the trends reveal that it is relatively constant, around 3.15$\times$10e5 number of requests per day in average for the entire city.

\begin{figure}[H]
	\centering
	\includegraphics[width=\textwidth]{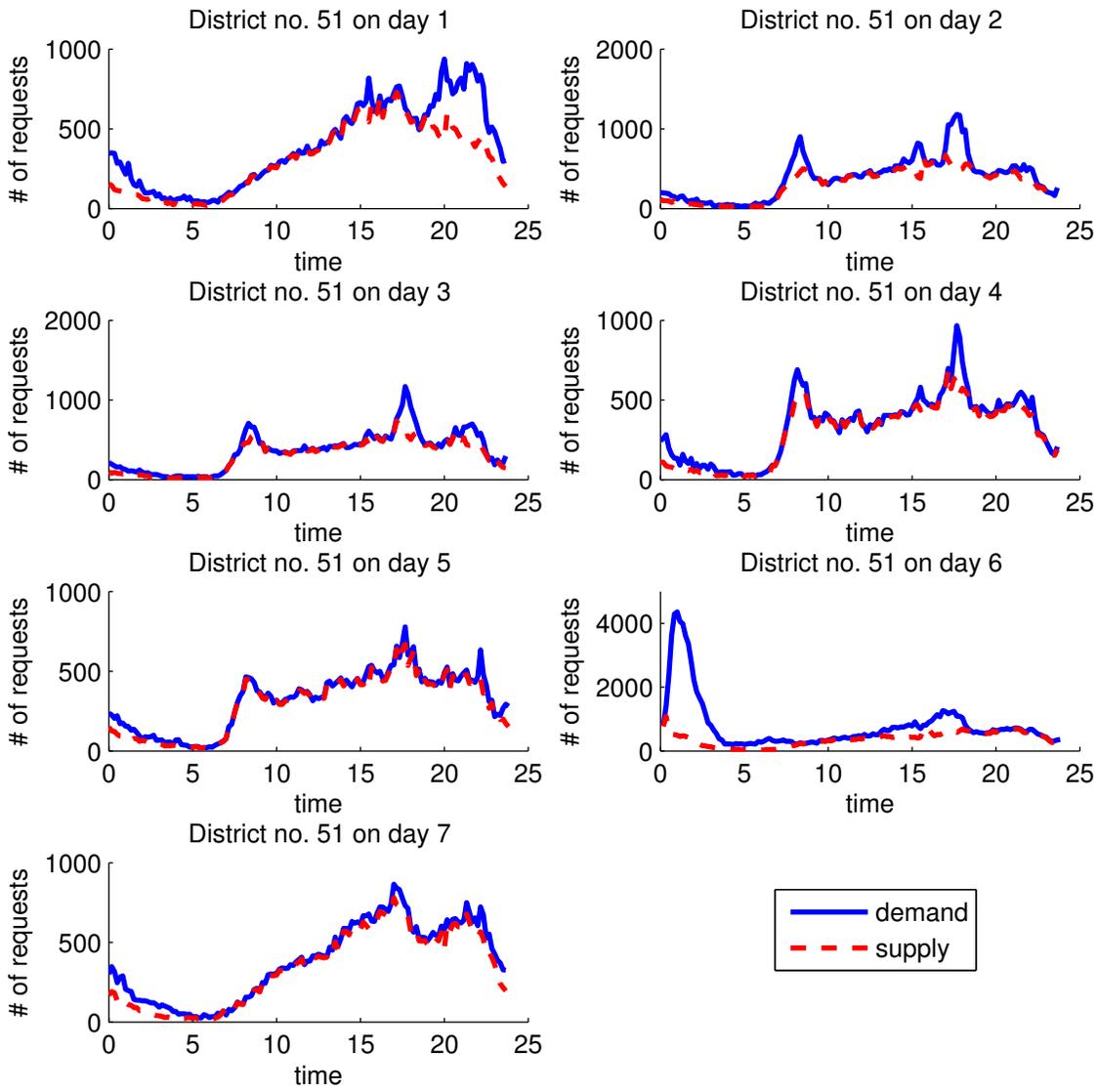}
	\caption{Evolution of the supply-demand according to district no.51 and days of week}
	\label{F:2}
\end{figure}

Figure \ref{F:2} presents the typical demand and supply patterns that can be observed within a district (e.g. 51 which presents the most important ride-hailing activity) and a day of week. The commuting patterns are clearly visible for weekdays with the peak-hours 7.00am-10.00am and 4.30pm-8.00pm. In Figure \ref{fig:3a}, we observe that the demand is not completely fulfilled during the peak-hours. The unmet demand can be illustrated by the two peaks appearing in the gap curve within the time intervals 7.00am-10.00am and 4.30pm-8.00pm. Figure \ref{fig:3b} describes the evolution of the average price including minimum, maximum and median. The average increases progressively from 1.00am with the first demand requests until 8.00am where the average price seems be more stable during the rest of the day.

\clearpage


\section{Modeling approaches}
\label{S:4}

The current section is dedicated to the description of the integrated framework which includes the features selection algorithm for selecting the most relevant variables and the machine learning algorithms for short-term demand forecasting. In Section \ref{S:4.1}, we present the RreliefF algorithm for variable selection and its main features. We also discuss why such a step is necessary before incorporating the explanatory variables as input into the different machine learning algorithms. In Sections \ref{S:4.2} and \ref{S:4-3}, we provide a detailed description of the different decision-trees based algorithms. The advantages and limitations of each algorithm are discussed. We also compare the specific methodological features characterizing each type of decision tree-based algorithm. Since these three algorithms belong to the same category, we will present them under a single section. Then, in Section \ref{S:4-4}, we describe the Gradient Boosting algorithm for regression adopted in the current study and how such an approach can be applied for forecasting the short-term ride-hailing demand. Finally, the ANN-based approach for regression is described in Section \ref{S:4.5} in order to be compared with the other methods.

\subsection{RreliefF algorithm}
\label{S:4.1}

In the current study, selection of the most relevant predictors or independent variables is performed using the RReliefF algorithm for regression. In doing so, the relative importance of predictors can me computed in addition to their ranking.The first version of the ReliefF algorithm was proposed by \citet{Kononenko1997} for classification only. To handle models for regression, a generalized version of the Relief algorithm has been developed \citep{Robnik_aikonja2003}. As outlined by \citet{Robnik_aikonja2003}, the algorithm is capable of capturing the conditional dependencies between attributes in order to provide a unique view on the relative importance of each single predictor with respect to the outcome or dependent variable. The RReliefF algorithm for regression is particularly useful in the preprocessing step for reducing the dimensionality of large datasets. A theoretical and empirical investigation of the Relief algorithms is presented in \citet{Robnik_aikonja2003} including a detailed discussion about their properties and methodological features. In this paper, the RReliefF algorithm will be briefly analyzed with an emphasis on the structure of the algorithm (see Algorithm \ref{Algo:0}) and its importance in the context of large datasets. In this regard, three important aspects should be highlighted:

\begin{enumerate}[label={\alph*)}]
	\item For large datasets, some predictors may be highly correlated which results in the mitigation of the general predictive capabilities of the fitted models. Incorporating redundant information into nonparametric models may affect or worsen the accuracy of the predicted outcomes \citep{chen2017understanding}.
	
	\item If the dimensionality of the dataset is reduced before training a model, the interpretation of the results can be easier and more efficient. Also the fitted models present more stability and robustness when a smaller set of relevant explanatory variables is selected. A smart selection of the predictors make the calibration process of base-learners or regression trees more efficient and less complex as non-useful predictors would lead to higher computational complexity and important run-time consumption \citep{chen2017understanding}.
	
	\item If the choice of the predictors is not optimized, then the models may end up with a training datasets that include a large amount of Non Attributed (NA) values. In this context, some observations could be removed from the dataset because of some irrelevant variables. And a decrease in the number of observations may be particularly problematic for small sample sizes. Note that in the current study, the size of the dataset is sufficiently large to ensure the training and the test phases \citep{saadi2016hidden,chen2017understanding}.
\end{enumerate}

\begin{algorithm}[H]
	\caption{Regression ReliefF Algorithm}
	\label{Algo:0}
	\begin{algorithmic}
		\STATE Input: for each training instance a vector of attribute values $x$ and predicted value $\tau(x)$ 
		\STATE Output: vector $w$ of estimations of the qualities of attributes
		\STATE Ensure that $N_{dC}=0$, $N_{dA}(A)=0$, $N_{dC-dA}(A)=0$ and $W(A)=0$;
		
		\FOR{$i=1$ to $m$}
		\STATE Randomly select instance $R_i$
		\STATE Select $k$ instances $I_j$ nearest to $R_i$
		
		\FOR{$j=1$ to $k$}
		\STATE $N_{dC}=N_{dC}+\Delta(\tau (.),R_i,I_j).d(i,j)$;
		
		\FOR{$A=1$ to $a$}
		\STATE $N_{dA}(A)=N_{dA}(A)+\Delta(A,R_i,I_j).d(i,j)$
		\STATE $N_{dC-dA}(A)=N_{dC-dA}(A)+\Delta(\tau (.),R_i,I_j).\Delta(A,R_i,I_j).d(i,j)$
		
		\ENDFOR
		\ENDFOR
		\ENDFOR
		
		\FOR{$A=1$ to $a$}
		\STATE $w(A)=\frac{N_{dC-dA}(A)}{N_{dC}}-\frac{N_{dA}(A)-N_{dC-dA}(A)}{m-N_{dC}}$
		\ENDFOR
		
	\end{algorithmic}
\end{algorithm}

Algorithm \ref{Algo:0} aims at updating the vector of weights $w$ of length $A$ where each element $w(a)$ is associated to an attribute $a$. The first step consists of a random selection of an instance $R_i$. Note that this operation will be repeated $m$ times where $m$ is a tuning parameter. Then, a set of $k$ nearest neighbors $I_j$ of $R_i$ are identified. The updating of vector $w$ depends on $N_{dC}$ (the different prediction value), $N_{dA}$ (the different attribute) and $N_{dC-dA}$ (the different predicted and different attribute simultaneously). For categorical attributes, the factor $\Delta(A,R_i,I_j)$ is 0 if $value(A,R_i)=value(A,I_j)$, 1 otherwise. For numerical attributes, the factor $\Delta(A,R_i,I_j)$ is given by $\frac{|value(A,R_i)-value(A,I_j)|}{max(A)-min(A)}$. The value of $d(i,j)$ is given by $\frac{d_*(i,j)}{\sum_{l=1}^{k}d*(i,l)}$ where $d_*(i,j)=\exp^{-(\frac{rank(R_i,I_j)}{\sigma})^2}$. In the later equation, $rank(R_i,I_j)$ is the rank of instance $I_j$ in a sequence of instances ordered according to the distance from $R_i$ where $\sigma$ is another tuning parameter. As a set of $k$ nearest neighbors is considered by Algorithm \ref{Algo:0}, $\Delta(A,R_i,I_j)$ should be interpreted as the sum of the distances between $R_i$ and the instances $I_j$ (Manhattan distance). In doing so, the vector $w$ is updated once all its corresponding components $N_{dC}$, $N_{dA}$ and $N_{dC-dA}$ are determined. For further details, one can refer to the work of \citet{Robnik_aikonja2003}.

\subsection{Decision Tree and Bagging}
\label{S:4.2}

\begin{figure}
	\centering
	\includegraphics[scale=0.6]{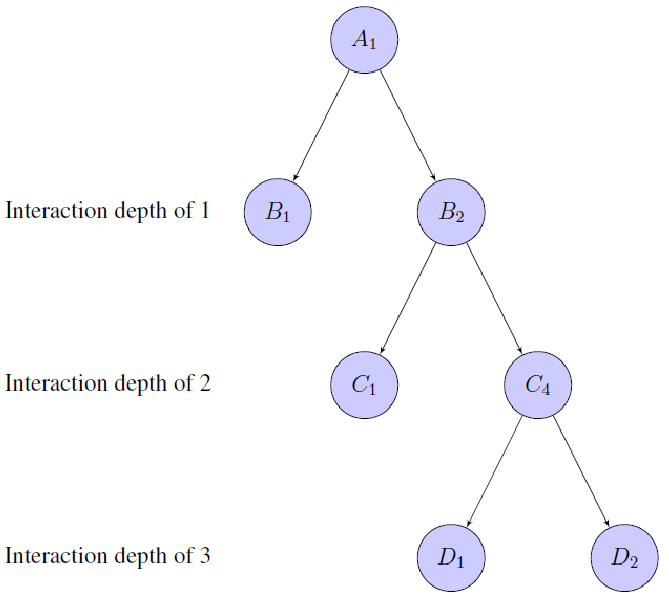}
	\caption{DT for regression predicting a single variable with a recursive procedure}
	\label{F:4}
\end{figure}

Decision Tree (DT) for regression is a predictive model capable of estimating a continuous (or categorical) outcome based on a set of explanatory variables. DT captures the decision making structure associated to a dataset. Fundamentally, the structure of a DT corresponds to a set of branches connected with nodes (see Figure \ref{F:4}). In practice, DT uses a recursive partitioning procedure to derive additional subsets. At each internal node of the tree, a test on one of the considered variables is realized. A branch represents the result of a test. According to the convention, when the expression at a node is true then the DT follows the left branch otherwise it follows the right branch. In doing so, the recursion can be realized efficiently while making the visualization of the tree and its interpretation easier. In mathematical terms, a training dataset has the form $(\textbf{x}_i,y_i)_{i=1}^{M}=(x_{1i},x_{2i},...,x_{ni},y_i)_{i=1}^{M}$ where $\textbf{x}_i$ represents the vector of the explanatory variables, $i$ the $i^{th}$ row observation and $\textbf{y}$ the targets or the dependent variable. In the context of the current study, $\textbf{y}$ will represent the number of requests according to space, time and under specific traffic, weather and pricing conditions.

In the current study, we also propose to calibrate an ensemble learning approach based on "bootstrap aggregating" or "bagging" strategy in order to investigate whether the prediction accuracy can be improved and to what extent. Fundamentally, the Bagged Decision Trees (BDT) include a wide range of grown trees in order to better capture the decision making inherent to the dataset. Bootstrap samples are extracted from the training dataset by sampling with replacement. Then, based on each single bootstrap sample, a decision tree is calibrated to capture the characteristics of its corresponding sample. The bootstrap samples are randomly extracted. The outcome of the BDT is determined by averaging the outcomes of all the decision trees according to Algorithm \ref{Algo:2} \citep{Breiman1996}. Bagging strategy provides more stability to the BDT. The bagging algorithm applied in the current study is defined in Algorithm \ref{Algo:1}.

\begin{algorithm}[H]
	\caption{Bootstrap aggregating "bagging" algorithm}
	\label{Algo:1}
	\begin{algorithmic}
		\STATE Ensure that $D=\varnothing$
		\STATE Set number of trees to be trained $n$
		\FOR{$i=1$ to $n$}
		\STATE Take a bootstrap sample $BS_k$ from $BS$, the training set
		\STATE Build a decision tree $D_k$ based on $BS_k$ (sub-training set)
		\STATE Add the calibrated decision tree to the ensemble $D$ such that $D=D\cup D_k$
		\ENDFOR
		\STATE Return $D$
	\end{algorithmic}
\end{algorithm}

\begin{algorithm}[H]
	\caption{Regression phase}
	\label{Algo:2}
	\begin{algorithmic}
		\STATE Ensure that the vector of outcomes $R=\varnothing$
		\STATE Given an input $x$
		\FOR{$i=1$ to $n$}
		\STATE Run the decision tree $D_i$
		\STATE Add the outcome $R_i$ of $D_i$ to $R$: $R=R\cup D_i$
		\ENDFOR
		\STATE Compute the average (or weighted average) of $R$
		\STATE Return $R$
	\end{algorithmic}
\end{algorithm}

It has been demonstrated in several studies that the bagging strategy improves the prediction accuracy \citep{Breiman1996}. In the context of unstable classifiers such as decision trees, the averaging over the bootstrap samples mitigates the error stemming from the variance.

\subsection{Random Forest}
\label{S:4-3}

Random Forest algorithm is also an ensemble learning method based on the combination of the outcomes of several decision trees. The classification or regression trees are trained on an independent random sampling of bootstraps characterized by the same distribution for all the trees \citep{breiman2001random}. In addition to the row observations bagging from the original dataset explained in Section \ref{S:4.2}, Random Forest (RF) algorithm randomly selects a subset of predictors for performing the decision splits \citep{breiman2001random}. Thus, it can be assimilated to a "predictor bagging" procedure or random subspace of predictors. In practice, the recommended number of random predictors for subspace sampling is the rounded down value of $\frac{n_p}{3}$ where $n_p$ is the number of predictors within the original dataset. Besides, more general predictive capabilities can be achieved for larger number of trees. As outlined by \citet{friedman2001elements}, the quality of each single tree of the forest and their correlation influence the generalization of the error.

\subsection{Boosting}
\label{S:4-4}

The Gradient Boosted Decision Trees (GBDT) is a powerful nonparametric statistical learning approach for classification and regression based on a sequential addition of weak learners. In the current study, weak learners correspond to regression trees as the on-demand ride-hailing is a continuous outcome. The GBDT model is characterized by an important flexibility as various loss functions can be used and different learning functions can be used. Also, compared to other conventional approaches, e.g. Support Vector Machines (SVM) or Artificial Neural Networks (ANN), GBDT models are relatively fast, robust and show competitive performance. Indeed, nonsensical predictions are extremely rare and it generally produces better estimates than simpler techniques. In addition to the previous advantages, GBDT explicitly handle NA values and do not require any normalization of the input predictors on the contrary of SVM for instance. While RF algorithm handles a maximum of 32 factor levels, GBDT can reach up to 1024 levels. With respect to the predictors, GBDT is relatively insensitive to eventual variable correlations and it is capable of handling a high number of predictors. The detailed sequences of the GBDT are presented in Algorithm \ref{Algo:3}.

\begin{algorithm}[H]
	\caption{Gradient Boosting Algorithm}
	\label{Algo:3}
	\begin{algorithmic}
		\STATE Input data $(\textbf{x}_i,y_i)_{i=1}^{N}$ where $\textbf{x}_i$ is the $i^{th}$ row observation
		\STATE Set number of iterations $M$
		\STATE Set a shrinkage parameter $\beta$
		\STATE Select an appropriate differentiable Loss Function, i.e. $\gamma(\textbf{y},F(\textbf{x}))=\frac{1}{2}(\textbf{y}-F(\textbf{x}))^2$
		\STATE Select an appropriate Base Learner Model $h(\textbf{x},\theta)$, i.e. a regression tree
		\STATE Initialize $F_0(\textbf{x})$ with a constant, i.e. $F_0(\textbf{x})= \operatorname{arg\,min}_{\gamma} \sum_{i=1}^{N} \gamma (\textbf{y},\beta)$
		\FOR{$m=1$ to $M$}
		\FOR{$i=1$ to $N$}
		\STATE Compute the negative gradient $g_{mi}(\textbf{x})=-\frac{\partial \gamma(\textbf{y},F(\textbf{x}_i))}{\partial F(\textbf{x}_i)}\Big|_{F(\textbf{x}_i)=F_{i-1}(\textbf{x}_i)}$
		\ENDFOR
		\STATE Fit the Base Learner Function, i.e. regression decision tree, $h(\textbf{x},\theta_m)$ to the residuals vector $\textbf{g}_m=(g_{m1},...,g_{mN})$ by using the following training set $(\textbf{x}_i,g_{mi})_{i=1}^{N}$
		\STATE Compute the best gradient descent step-size according to the following one dimensional optimization problem $\beta_m= \operatorname{arg\,min}_{\beta} \sum_{i=1}^{N} \gamma (\textbf{y},F_{m-1}(\textbf{x}_i)+\beta h_m(\textbf{x}_i))$
		\STATE Update the function estimate $F_m(\textbf{x})=F_{m-1}(\textbf{x})+\beta_m h_m(\textbf{x})$
		\ENDFOR
		\STATE Return $F(\textbf{x})$
	\end{algorithmic}
\end{algorithm}

GBDT includes an important feature that is based on a loss function. In the current study, the Gaussian loss (or the squared-error $L_2$) function $\gamma(\textbf{y},F(\textbf{x}))=\frac{1}{2}(\textbf{y}-F(\textbf{x}))^2$ will be tested. As the derivative of $\gamma(\textbf{y},F(\textbf{x}))$ with respect to $F(\textbf{x})$ is $f-y$, it means that GBDT performs a residual fitting. Therefore, vector $\textbf{g}_m$ includes all the residuals computed across all the row observations $i$ at each iteration $m$. The choice of an appropriate loss function is of great importance as in our application, some outliers corresponding to important demand levels, need to be characterized and correctly predicted while maintaining general predictive capabilities. In this regard, different regularization aspects are considered within the current framework to avoid that GBDT over-fit the training dataset. Sub-sampling is the simplest procedure for model generalization. 

As mentioned in Algorithm \ref{Algo:3}, a weak learner, i.e. a regression tree in our case, is trained based on the training set. Instead of considering all the observations, only a random part of the training set without replacement is selected for training. It results in two advantages: high computer run-time is mitigated and some randomness is incorporated into the overall GBDT model. In the context of large datasets, sub-sampling is particularly interesting as the training dataset can be fitted by taking lower fractions of the original training dataset. When the number of row observations of the training dataset is reasonable, a bag fraction of 50\% provides a good predictive accuracy in many applications \citep{natekin2013gradient}. In the case of data from ride-hailing demand, we will show that a slightly higher value provides better estimates. In practice, a trade-off is found between the bag fraction and the number of iterations. Indeed, if the bootstrap sample is too small then higher number of iterations are necessary to reach a good predictive accuracy while equivalent results can be achieved with a smaller number of iterations and higher bootstrap sample size.

The shrinkage is another fundamental parameter in GBDT models \citep{friedman2001greedy}. This parameter mitigates the effects of each new fitted regression tree. Instead of adjusting the objective function with few important successive regression tree outputs, lower contributions are captured but for a wider set of regression trees or iterations. In this regard, if one of the base learners turns out to be badly calibrated then its negative impact is covered by the outcomes of the other ones. The shrinkage parameter $\beta$ varies between 0 and 1. If $\beta$ is small then more iterations $I$ are needed to reach a satisfactory level of accuracy. For high dimensional datasets, values in-between 0.01 and 0.1 provides good predictions in general.

The final technique for regularization is the early stopping. As highlighted by \citet{zhang2005boosting}, choosing the highest possible number of iterations does not automatically guarantee that the results will be better. As it will be investigated in Section \ref{S:5}, for specific values of $\beta$, the error curve takes a concave up shape with a re-increase of the error after reaching the minimum. Indeed, by increasing continuously the number of iterations, this means that GBDT tends to over-fit the training dataset while mitigating its general predictive capabilities with respect to the validation dataset.

Finally, another important feature of the GBDT is the choice of the base learner, i.e. regression tree. The quality of the regression tree depends on different characteristics. The interaction depth is of great importance or the maximum tree depth. An optimal choice between stumped and deep regression trees has an important impact on the final outcome accuracy.

\subsection{Artificial Neural Network}
\label{S:4.5}

Artificial Neural Networks (ANN) models are based on the operating principle of the human brain \citep{graupe2013principles}. Similar to the decision trees-based algorithms, our ANN-based approach is also an ensemble learning technique. The neurons send activation signals downstream, i.e. forward propagated, towards the output node and the error is backward propagated using a stochastic gradient decent algorithm. The most popular ANN-based methods use more elaborated functions, calibration and tuning parameters, e.g. Dropout, ConvNets, and long-short term memory (LSTM) networks \citep{srivastava2014dropout,krizhevsky2012imagenet,graves2013hybrid}. Recent unsupervised learning methods have allowed for algorithms to be trained on very large unlabeled datasets. 

From a methodological point of view, each independent variable (neuron) can affect the class under through a series of weighted functions and biases. Let us denote a neuron as $x$, the connection weights as $w$ and the activation function as $F$, then the artificial neuron network function can be formulated as follows $f_j(x)= F(\sum_i w_{ij} g_i(x) + b_j)$, where $g_i(x)$ is the output from the previous node. Stochastic gradient decent is used to update the weights moving it closer to the optimal point with respect to the loss function and updates are performed on each mini-batch iteration: $w_{ij (new)} = w_{ij(old)} - \alpha \frac{\partial \mathcal{L}(w_{ij})}{\partial w_{ij}}$, where $\alpha$ is the learning rate and $\frac{\partial \mathcal{L}(w_{ij})}{\partial w_{ij}}$ is the derivative of the loss function with respect to the parameters. Learning methods compute the optimal weights given a set of "hyper parameters" such as the number of neurons, number of layers and learning rate. The degree of complexity can be tuned accordingly while training on the same set of data. The optimal model is guided by finding the set of functions and hyper parameters that gives the best validation results.

\section{Results and discussion}
\label{S:5}

In this section, the results obtained from each approach are presented and individually compared with respect to the validation dataset. The machine learning approaches outlined in Section \ref{S:4} are systematically compared using various statistics to better assess which technique performs better and under what assumptions. The first step of the modeling process consists of preprocessing the explanatory variables using the RreliefF algorithm in accordance with Section \ref{S:4.1}. The dataset that describes the evolution of the demand for ride-hailing during 21 days, from the 1st to the 21st of January 2016, is provided by the on-demand ride-hailing service company DiDi for an anonymized city of China. The 66 districts of the study area are referred by using indexes ranging from 1 to 66. Table \ref{T:0} presents all the variables considered within the modeling process.

\subsection{Variable importance}
\label{S:5.1}
The parameter $k$ in RreliefF algorithm that corresponds to the $k$-nearest neighbor(s) $k$ has an important influence on the final RreliefF algorithm outcome as highlighted in Section \ref{S:4.1}. Thus, its value should be the most optimal to ensure that the results are correct. As we are handling a large dataset, testing the algorithm for several values of $k$ would add more computational complexity with an increase in the computer run-time. Therefore, to be more conservative, we decided to set the value of $k$ at the maximum, i.e. the total number of observations, 199,584. Setting up such value of $k$ is possible thanks to the assumption of weight importance stability. Indeed, after a certain number of $k$ iterations, the weight importance of each single variable does not vary anymore. The results presented in Table \ref{T:var_relief} reveal that the variables district-id, fare level, day of week and weather have an effect on the demand. Note that the supply and gap related variables were excluded, as their effects are already captured by the fare level. In the context of ride-hailing, the supply is rather demand-driven, while the gap is a linear combination of the demand and the supply variables. The rest of the variables will be considered as input by the RreliefF algorithm.

As presented in Table \ref{T:var_relief}, time of day has an important effect. The demand is particularly high during peak hours than during off-peak hours. Intuitively, one can say that there is a clear relationship between time of day and the within-day demand. This relationship has been successfully captured by the RreliefF algorithm with a ranking of four. Also, the daily-demand seems to be sensitive to variables related to ride fares, i.e. average and maximum prices. They have an equivalent contribution (0.036 and 0.037) with respect to the explanation of the demand while the other ride prices are a bit less important (0.022 and 0.024). The variables related to weather conditions showed lower significance, but we kept them as their impact on the demand is not negligible. We are characterizing the temporal dynamics of the demand across days, so dow variable is preserved as well. The small weight associated to dow can be explained by the fact that the demand patterns as a function of day are relatively similar across the spatial districts and from day to day although minor differences can be captured between week-days and week-ends. 

Regarding the variables describing the traffic conditions, the results show that the LoS at the global scale (city) have a lower impact on the demand than those at the local scale, which makes sense, as rider is only concerned with the local congestion that they experience themselves. At the local scale (district), the LoS from 1 to 2 have the largest weights for explaining the demand. In contrast, the tj-level-4-m10 has a very small impact.

Note that the RreliefF algorithm didn't detect any negative weights, thus it means that all the variables have a relative impact on the demand. Thus, the variables only differs in terms of intensity. If few of the lowest-weighted variables variables are dropped, the effects would be very minor with respect to the predictive capabilities of the algorithms.

Contrary to the findings of \citet{chen2017understanding} for ride-splitting, the outcome of the RreliefF algorihtm reveals that, in the context of ride-hailing demand prediction, all the variables should be taken into account as they are characterized by positive weights. Inversely, variables associated to null or negative weights would have been dropped, which is not the case here.

\begin{landscape}
	\begin{center}
		\begin{table}[H]
			\centering
			\begin{tabular}{|l|l|l|l|l|l|r|r|r|r|}
				\hline
				ID	& Variable & Description & Type &	Unit &	Range &	Mean &	S.D. &	Median& P. (*) \\
				\hline
				\hline
				\multicolumn{10}{|l|}{\textit{Space-Time}}\\
				\hline
				
				1	& District-id & District ID &	Cat. & N.A. & $\left[1;66\right]$ &- &- &-&$\times$ \\
				2	&Hour& Hour	&Cont.	&Hour	& $\left[0;23\right]$&-	&- &-&$\times$	\\
				3	&Minute&	Minute&Cont.&	Minutes&	$\left[0;59\right]$&- &- &-& $\times$	\\
				4	&Dow&	Day of week&Cat.	&Day	&$\left[1;7\right]$	&- &- &-	&$\times$\\
				
				\hline
				\multicolumn{10}{|l|}{\textit{Ride Requests (10 minutes aggregation)}}\\
				\hline
				
				5	&Demand-m10& \# of ride requests &Cont.&	-	&$\left[0;4362\right]$ &42.79&103.65&7.00&	$\times$\\
				6	&Supply-m10	& \# of fulfilled ride requests &Cont.	& -&	$\left[0;1084\right]$ &35.19&75.19&6.00&-\\
				7	&Gap	&Difference between Dem. and Sup.&Cont.	& -	& $\left[0;3872\right]$ &7.60&45.23&1.00&-\\
				\hline
				\multicolumn{10}{|l|}{\textit{Pricing}}\\
				\hline
				8	&Price-avg	& mean ride cost &Cont.&	M.U.(*)&	$\left[0;489\right]$&17.74&16.86&16.38&			$\times$\\
				9	&Price-median&	median ride cost&Cont.	&M.U.	&$\left[0;489\right]$	&14.08&15.38&12.00&	$\times$\\	
				10	&Price-min	&minimum ride cost&Cont.	&M.U.&	$\left[0;489\right]$	&7.08&12.91&4.00&$\times$\\
				11	&Price-max	&maximum ride cost&Cont.	&M.U.&	$\left[0;1731\right]$	&51.10&53.14&42.00&$\times$	\\
				12	&Destinations&	number of unique destinations&Cont.	&NA(*)&	$\left[1;63\right]$	&- &- &-&	$\times$\\
				\hline
				\multicolumn{10}{|l|}{\textit{Experienced Levels of Service (LoS) with 10 minute aggregation}}\\
				\hline
				
				13	&Tj-level-1-m10 (**)& Percentage of vehicles experiencing	LoS 1 & Cont.&		&$\left[0;1\right]$	&0.78&0.16&0.81	&	$\times$\\
				14	&Tj-level-2-m10&Per. veh. experiencing	LoS 2&Cont.&		&$\left[0;1\right]$	&0.12&0.07&0.12	&	$\times$\\
				15	&Tj-level-3-m10&Per. veh. experiencing	LoS 3&Cont.&		&$\left[0;1\right]$	&0.04&0.03&0.04	&$\times	$\\
				16	&Tj-level-4-m10&Per. veh. experiencing	LoS 4&Cont.&		&$\left[0;1\right]$	&0.03&0.02&0.02	&	$\times	$\\
				17	&Tj-global-1-m10(***)&Per. veh. experiencing	LoS 1&Cont.&		&$\left[0;0.91\right]$	&0.80&0.11&0.81&		$\times$\\
				18	&Tj-global-2-m10&Per. veh. experiencing	LoS 2&Cont.&		&$\left[0;21\right]$	&0.12&0.03&0.13	&	$\times$\\
				19	&Tj-global-3-m10&Per. veh. experiencing	LoS 3&Cont.&		&$\left[0;0.08\right]$	&0.04&0.01&0.04	&	$\times$\\
				20	&Tj-global-4-m10&Per. veh. experiencing	LoS 4&Cont.&		&$\left[0;0.05\right]$	&0.03&0.01&0.03	&	$\times$\\
				\hline
				\multicolumn{10}{|l|}{\textit{Weather}}\\
				\hline
				
				21	&Weather-m180&	Weather index (every 180 minute)&Cat.	&	&$\left[0;9\right]$	&- &- &-&$\times$\\
				22	&Temperature-m180&	temperature (every 180 minute)&Cat.	&	&$\left[0;19\right]$	&6.15 &3.76 &6.00&$\times$\\
				23	&PM2.5-m180&	PM2.5 level (every 180 minute)&Cat.	&	&$\left[0;264.9\right]$	&116.02&55.66&113.70&$\times$\\
				\hline
				\multicolumn{10}{|l|}{(*): Incorporated predictors are indicated with a cross according to the RreliefF algorithm}\\
				\multicolumn{10}{|l|}{M.U.:Monetary Unit, NA: Non-Attributed, (**): At district scale, (***): At city scale}\\
				\hline
			\end{tabular}
			\caption{Comparison between the predicted and observed patterns derived from the test dataset}
			\label{T:0}
		\end{table}
	\end{center}
\end{landscape}



\begin{table}[H]
	\centering
	\begin{tabular}{|l|c|}
		\hline
		Variables & Weight importance \\
		\hline
		destinations    & 0.459\\
		district-id     &	0.171\\
		tj-level-1-m10  &	0.134\\
		tj-level-2-m10  & 0.101\\
		time of day 		  & 0.047\\
		tj-global-level-2-m10	& 0.039\\
		price-avg		  & 0.037\\
		price-max		  & 0.036\\
		tj-global-level-3-m10	& 0.024\\
		price-median	& 0.024\\
		price-min	& 0.022\\
		tj-global-level-4-m10	& 0.019\\
		tj-global-level-1-m10	& 0.019\\
		tj-level-3-m10	& 0.013\\
		weather-m180	& 0.010\\
		dow	& 0.009\\
		PM2.5-m180	& 0.008\\
		temperature-m180	& 0.007\\
		tj-level-4-m10	& 0.003\\
		\hline
	\end{tabular}
	\caption{Ranking of the variables according to their weight importance}
	\label{T:var_relief}
\end{table}

\subsection{Decision Tree and Bagging}
\label{S:5.2}

With respect to the individual regression Decision Tree (DT), the calibration is quite straightforward. All the predictors are sampled. The minimum number of leaf node observations is 1 and the minimum number of branch node observations is 10. These are generally good default values. While the calibration of an individual regression decision tree model is not complex, such model cannot guarantee good prediction accuracy and would tend to over-fit the training dataset. In this regard, ensemble decision trees are more stable in case of transferability towards other datasets, i.e. validation dataset. This phenomenon is illustrated in the following paragraphs. In this comparative study, the single tree is considered as "base case" to better assess the improvements brought by more sophisticated algorithms.

In contrast, to set up the Bagged Decision Tree (BDT) model with a good prediction accuracy level, various parameters need to be optimally tuned. The first parameter corresponds to $n$, the number of trees included within the ensemble learning. According to \citet{Breiman1996}, the Mean Square Error (MSE) decreases when the number of trees used to characterize the dataset increases. In our case study, different values of $n$ have been tested according to the following sequence, for example: 10, 50, 100. As the error becomes stationary above a certain number of trees, it is no more optimal to indefinitely increase $n$. Besides, the number of decision trees is particularly related to the bagging strategy, since it combines the results of several decision trees using bootstrap samples of the training dataset. In the current model, 100 regression trees were largely found to be sufficient to characterize the whole training dataset and achieve accurate predictions. Note that for every regression tree, all the predictors are sampled for the decision splits.

\subsection{Random Forest}
\label{S:5.4}

For Random Forest (RF) algorithm, we also chose the same number of trees as on the BDT algorithm as here also the MSE reached its stability level. In addition to BDT, RF algorithm includes the random subspace selection strategy which consists of a random selection of the number of predictors for the decision splits. The number of sampled variables $\delta$ is lower than the total number of predictors included within the model. By default, the value is equal to one third of the total number of predictors for regression. However, in the current application, the results reveal that better prediction accuracy can be achieved if the number of variables to sample is higher. In contrast, it should not be too high, otherwise the positive effects of the random subspace selection on the error is mitigated. In which case, this would amount to calibrating a BDT by systematically sampling all the predictors. By testing different values of $\delta={33,50,66,83,99}$, we can conclude that with $\delta=50$, we obtain the lowest RMSE, i.e 23.5. Table \ref{T:RF_var} presents the other values of $\delta$.

\begin{table}[H]
	\centering
	\begin{tabular}{|l|c|}
		\hline
		$\delta$ & RMSE \\
		\hline
		99 (Max)	& 24.29\\
		83&	23.83\\
		66 &	23.56\\
		50 &	\textbf{23.50}\\
		33 \citep{breiman2001random}& 23.65\\
		\hline
	\end{tabular}
	\caption{Variation of the RMSE with respect to the number of sampled variables for RF algorithm}
	\label{T:RF_var}
\end{table}

\subsection{Comparison between DT, BDT and RF}

\begin{figure}[H]
	\centering\includegraphics[width=\textwidth]{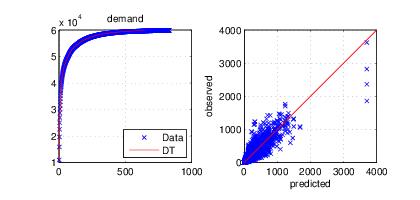}
	\caption{Comparison between the predicted and observed demand patterns based on DT}
	\label{sdt}
\end{figure}

\begin{figure}[H]
	\centering\includegraphics[width=\textwidth]{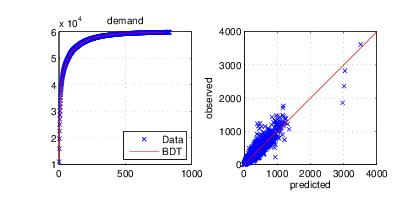}
	\caption{Comparison between the predicted and observed demand patterns based on BDT}
	\label{bdt}
\end{figure}

\begin{figure}[H]
	\centering\includegraphics[width=\textwidth]{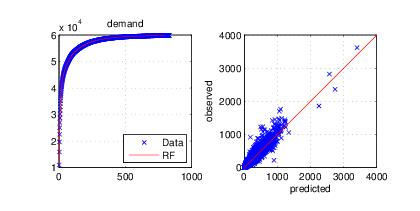}
	\caption{Comparison between the predicted and observed demand patterns based on RF}
	\label{rf}
\end{figure}

The results stemming from Figures \ref{sdt}-\ref{bdt}-\ref{rf} present the comparison between the predicted and observed space-time ride-hailing demand. This is done on the basis of the cumulative distribution and the joint distribution. Regarding the cumulative distributions, no matter the approach, the fit between the predicted and observed patterns are very similar. In practice, the predicted and observed values have been classified according to frequencies followed by a cumulative summation of the values. One can consider that from an aggregate perspective, the prediction capabilities of the three methods are quite accurate. However, if the results are compared on the basis of the joint distributions, DT is more affected by the spreading effects as illustrated in Figure \ref{sdt}. DT tends to over-fit the training dataset, which mitigates the prediction accuracy when the predicted demand is cross-validated. The outliers (high demand patterns) are successfully captured by RF (Figure \ref{rf}). In contrast, two outliers are badly predicted if BDT is applied (Figure \ref{bdt}) while DT captures only one of them.

The spread can also be assessed by computing the RMSE (Root Mean Square Error) by computing the following formula: $RMSE=\sqrt[]{E((y_{predicted}-y_{observed})^2)}$ where $E$ is the mean. The $RMSE$ is not normalized because the vectors that are being predicted contain a large number of zeros. In the normalized formulation of the $RMSE$, a division is indeed required such that $\frac{y_{predicted}-y_{observed}}{y_{observed}}$.

As BDT and RF have almost the same algorithmic structure, so now let us compare the contribution of their respective features--where DT would be the reference model. In Table \ref{T:1}, the predicted and observed space-time demand patterns are compared using different statistics. RF approach provides the best prediction accuracy with a RMSE of 23.50 and an R-square of 0.95 for $\delta$=50. As expected, DT has the higher RMSE of 33.55, although it is able to predict a higher maximum value of 3,701. Based on the RMSE, we can observe that including the bagging strategy within BDT mitigates the error by 38\%. In addition to bagging, the random sampling of variables at decision splits included within the RF-based approach mitigates the error by 3.25\% only with respect to BDT. 

In order to assess the stability of the different approaches when they are tested on independent datasets, the metrics are also estimated for the aim of comparison in the training dataset. In doing so, the effects of over-fitting can be illustrated through the value differences between Tables \ref{T:1} and \ref{T:2}. If important changes occur, it means that the models are over-fitted with the training dataset, which leads to low prediction capabilities. However, if the changes are stable, it means that the models present good prediction capabilities with a limited dependence to the training dataset.

\begin{table}[H]
	\centering
	\begin{tabular}{|l|l|l|l|l|l|l|l|l|}
		\hline
		Method & R-square & slope & RMSE & Mean & SD & Median & Min & Max \\
		\hline
		DT	& 0.89	&0.95&	33.55&	42.21&	101.68&	8.0&	0&	3,701\\
		BDT&	0.94&	1.01&	24.29&	42.33&	97.79&	8.0&	0&	3,516\\
		RF ($\delta=50$)&	0.95&	1.02&	23.50&	42.21&	96.54&	8.0&	0&	3,407\\
		Observed dataset&	-&	-&	-&	42.38&	101.38&	7.0&	0&	3,620\\
		\hline
	\end{tabular}
	\caption{Comparison between the predicted and observed patterns derived from the test dataset}
	\label{T:1}
\end{table}

\begin{table}[H]
	\centering
	\begin{tabular}{|l|l|l|l|l|l|l|l|l|}
		\hline
		Method	&R-square	&slope&	RMSE&	Mean	&SD	&Median&	Min	&Max\\
		\hline
		DT	&0.98	&1.00	&14.02&	42.93	&103.68&	8.0	&0&	3,701\\
		BDT&	0.98	&1.02	&15.00	&42.90	&101.12	&8.0	&0&	3,527\\
		RF ($\delta=50$)	&0.98	&1.03	&15.47	&42.92	&100.30	&8.0	&0	&3,480\\
		Data	&-	&-	&-	&42.96	&104.61	&7.0	&0	&4,362\\
		\hline
	\end{tabular}
	\caption{Comparison between the predicted and observed patterns derived from the training dataset}
	\label{T:2}
\end{table}

In Table \ref{T:2}, as expected DT has the lowest RMSE with a very good fit in terms of R-square and slope. The reason is that DT tends to over-fit the training dataset. With respect to the corresponding RMSE of Table \ref{T:1}, the error increased by 136\%. In contrast, the RMSE values related to BDT and RF increased by 63.93\% and 51.91\% respectively. Thus, RF is capable of capturing the characteristics of the training dataset with the smallest level of over-fitting. As expected, one can conclude that RF is more stable than the other methods in terms of transferability.

\subsection{GBDT}
\label{sub:5.5}

\begin{figure}[H]
	\centering\includegraphics[width=\textwidth]{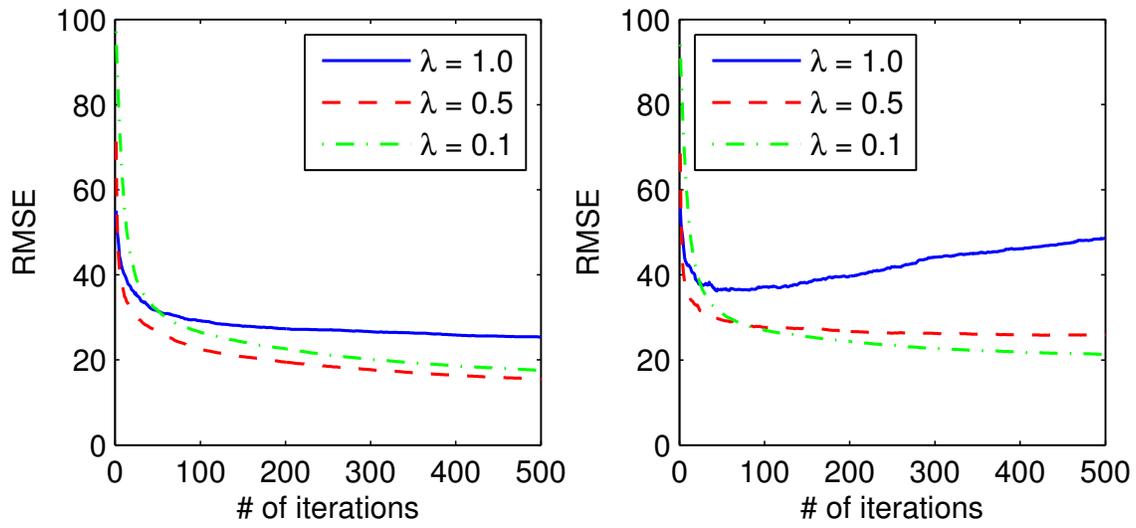}
	\caption{Error curves for GBM fitting on training and validation datasets}
	\label{error_curves_gbm}
\end{figure}

With respect to the calibration of the GBDT, several parameters have been calibrated successively to reach the most optimal values. As mentioned in Section \ref{S:4-4}, four important parameters influence significantly the accuracy of the predicted values:

\begin{enumerate}
	\item The number of trees $ntrees$: Like the two previous ensemble learning approaches (BDT and RF), the quality of the predicted values is improved when the number of trees or iterations is high. The number of trees is also the number of additive functions within the ensemble learning model. Of course, it is not interesting to go beyond a specific number of iterations since the error becomes stable. In the current ride-hailing application for demand estimation, the number of weak learners ranges from 1000 to 6000. One of the various advantages of the GBDT algorithm is that one can run the algorithm until a given value and then examine the outputs. Afterwards, starting from the current state, additional iterations can be added. This feature is particularly interesting for computer run-time reduction.
	
	\item The learning rate for shrinkage or step-size reduction parameter $\beta$ is also a key component within the GBDT algorithm. The default value is 1. Figure \ref{error_curves_gbm} presents the influence of the parameter for different models with respect to the training dataset (left graph) and the validation dataset (right graph). Regarding the validation dataset, if the learning rate for shrinkage is 1 then the minimum MSE is obtained at the near-beginning part of the curve. The curve tends to increase for higher number of iterations. In this context, the learning rate for shrinkage parameter should always be smaller to slow down the learning process and mitigate the risk of over-fitting. In \citet{friedman2001elements}, one can find that values ranging from 0.1 and 0.001 are considered to be the most adapted. However, for large datasets, i.e. more than 10,000 time-slots or observations, 0.1 is better. Indeed, our experiments reveal that smaller values, i.e. 0.01 or even 0.001, did not add substantial improvements to the model or the results tend even to be worse while increasing unnecessarily the computer run-time. Thus, a value of 0.1 has been fixed in the current GBDT model since a large dataset is handled.
	
	\item Regarding the learning functions or weak learners, they also have a substantial influence depending on their complexity. Generally, deeper trees are grown to reach better predictive accuracy. The simplest weak learner configuration is called stump (one interaction depth). The most complex is deep tree (highest possible interaction depth, e.i. number of variables). In the current case study, given the size and the complexity of the dataset, the tests revealed that deep trees are more adaptive. Thus the tree depth has been set to 19. Of course, intermediary values of interaction depth can be considered if necessary for particular case studies. Also another parameter related to the complexity of the trees can be taken into account to ensure that the number of observations at the leaf or terminal nodes is maintained higher or equal to a minimal value. Let us set this parameter to 10 according to the existing literature.
	
	\item In addition to the learning rate for shrinkage parameter, the bag fraction can also be incorporated with the GBDT to mitigate the effects of over-fitting and to ensure the generalization of the predictive models. A value of 0.5 works quite well for many applications. Given the complexity of the dataset, our experiments show that a bag fraction value of 0.7 provides more power to GBDT algorithm for capturing the spatio-temporal dynamics of the short-term demand with respect to the training dataset.
	
\end{enumerate}

\begin{figure}[H]
	\centering
	\begin{subfigure}[b]{0.45\textwidth}
		\includegraphics[width=\textwidth]{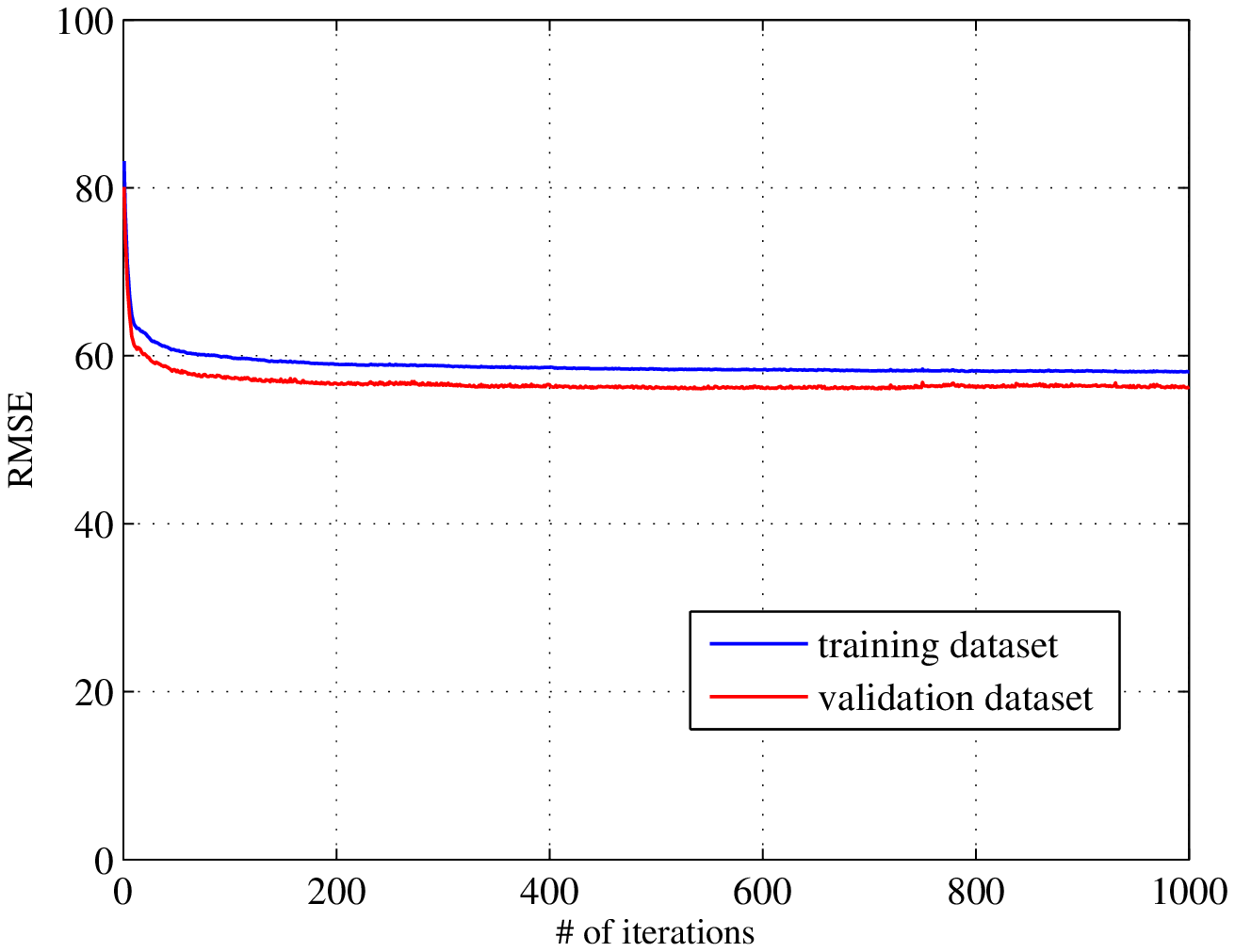}
		\caption{$\beta$=1, idepth=1, bag fraction=0.5, ntrees=1000}
		\label{fig:4a}
	\end{subfigure}
	\begin{subfigure}[b]{0.45\textwidth}
		\includegraphics[width=\textwidth]{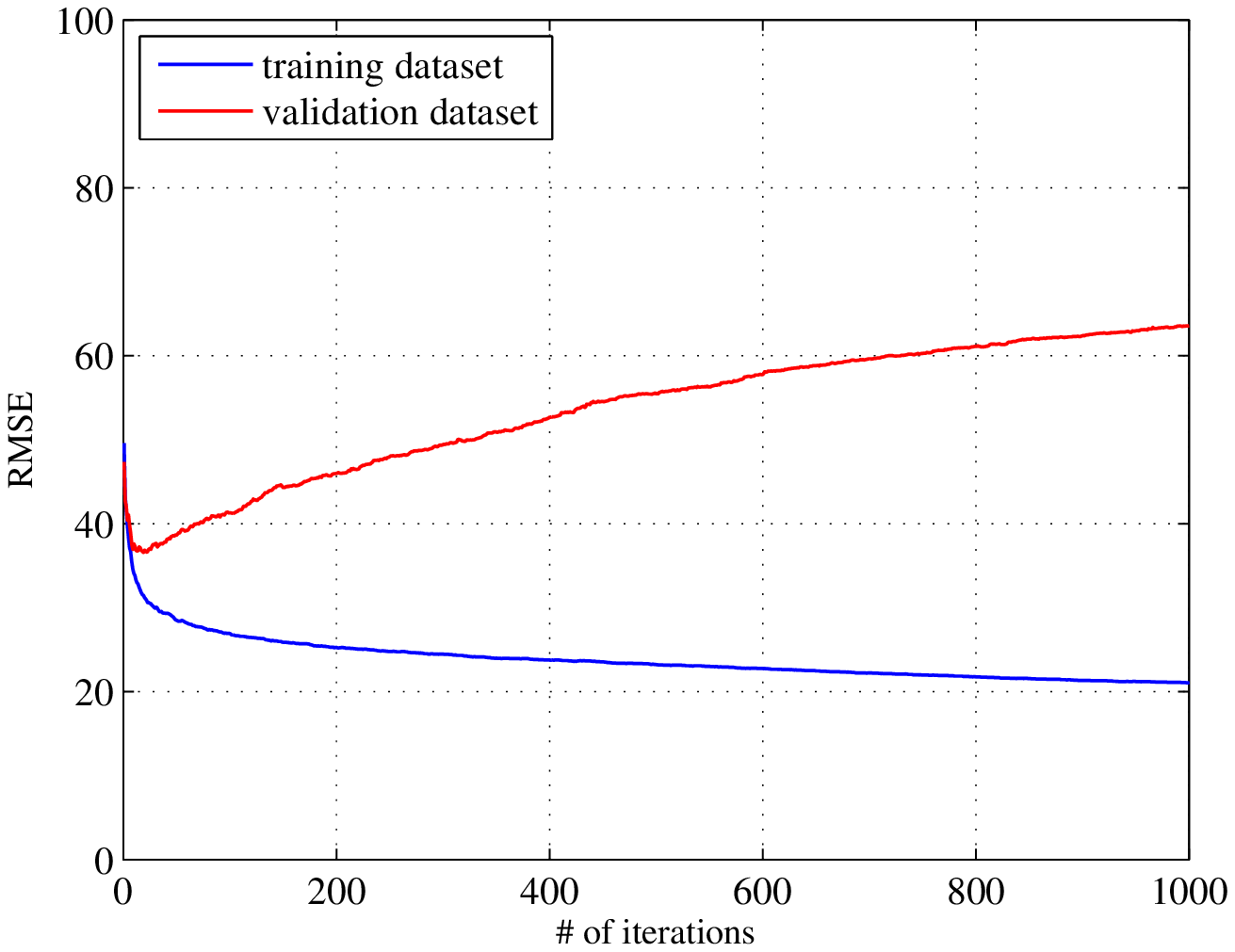}
		\caption{$\beta$=1, idepth=19, bag fraction=0.5, ntrees=1000}
		\label{fig:4b}
	\end{subfigure}
	\begin{subfigure}[b]{0.45\textwidth}
		\includegraphics[width=\textwidth]{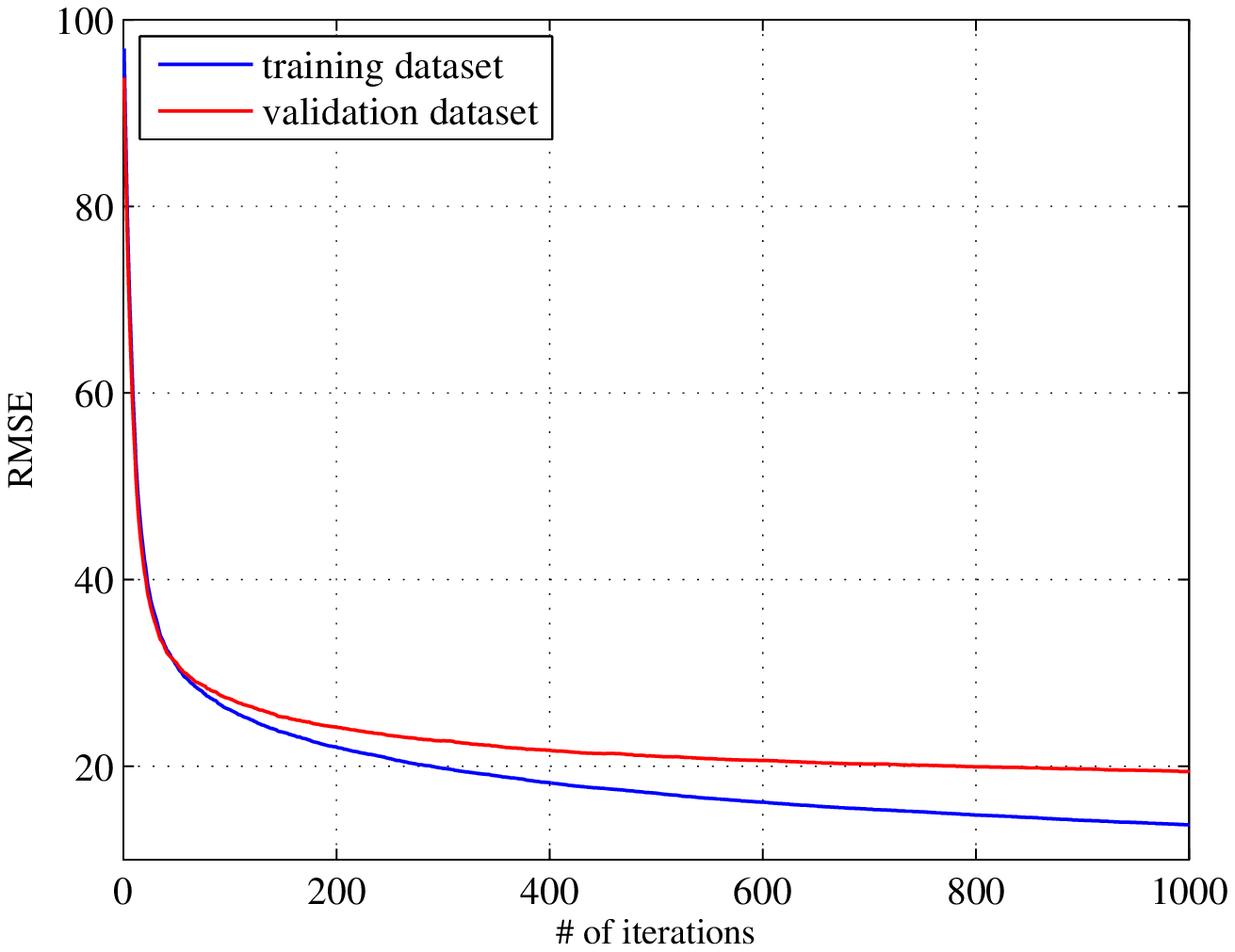}
		\caption{$\beta$=0.1, idepth=10, bag fraction=0.5, ntrees=1000}
		\label{fig:4c}
	\end{subfigure}
	\begin{subfigure}[b]{0.45\textwidth}
		\includegraphics[width=\textwidth]{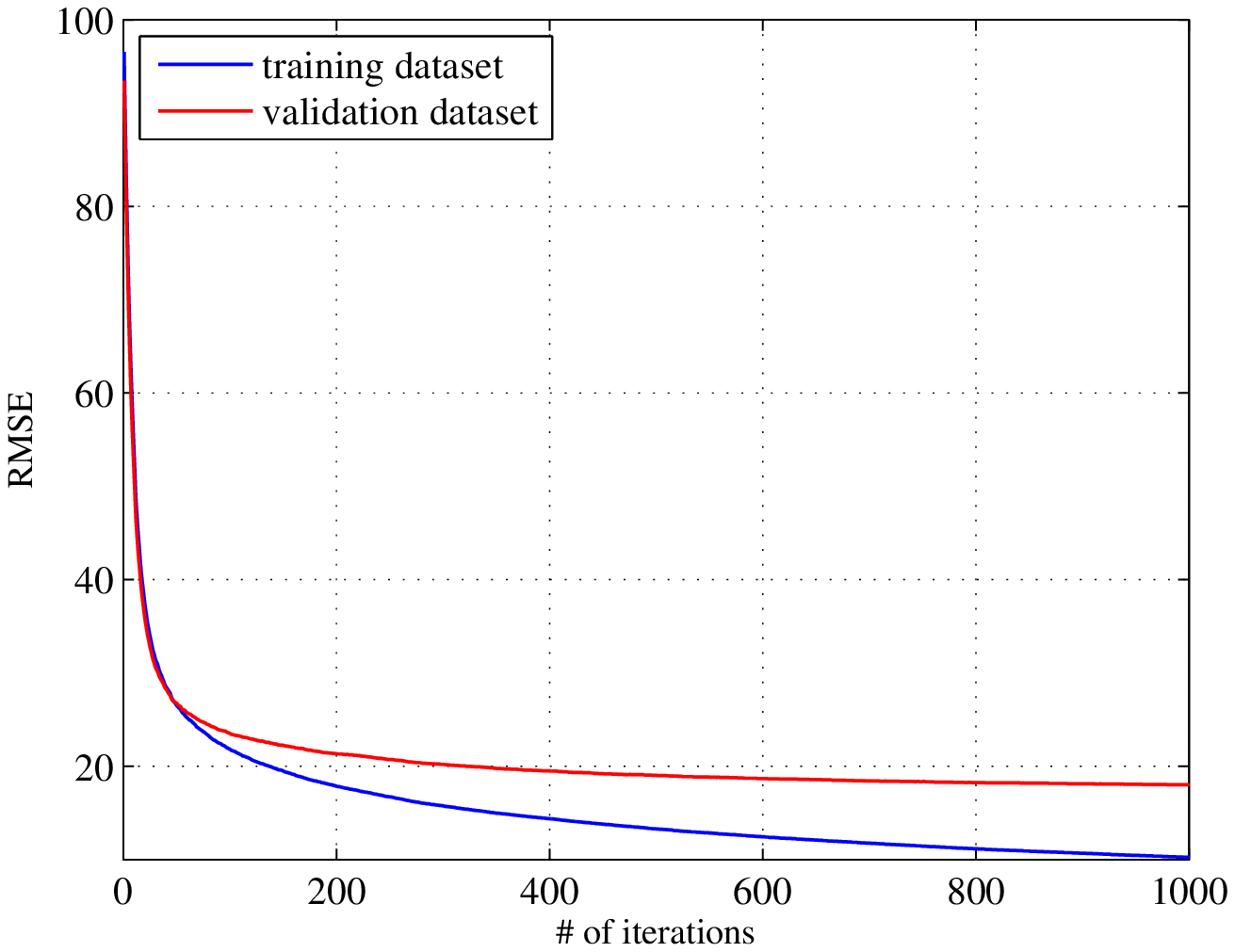}
		\caption{$\beta$=0.1, idepth=19, bag fraction=0.5, ntrees=1000}
		\label{fig:4d}
	\end{subfigure}
	\begin{subfigure}[b]{0.45\textwidth}
		\includegraphics[width=\textwidth]{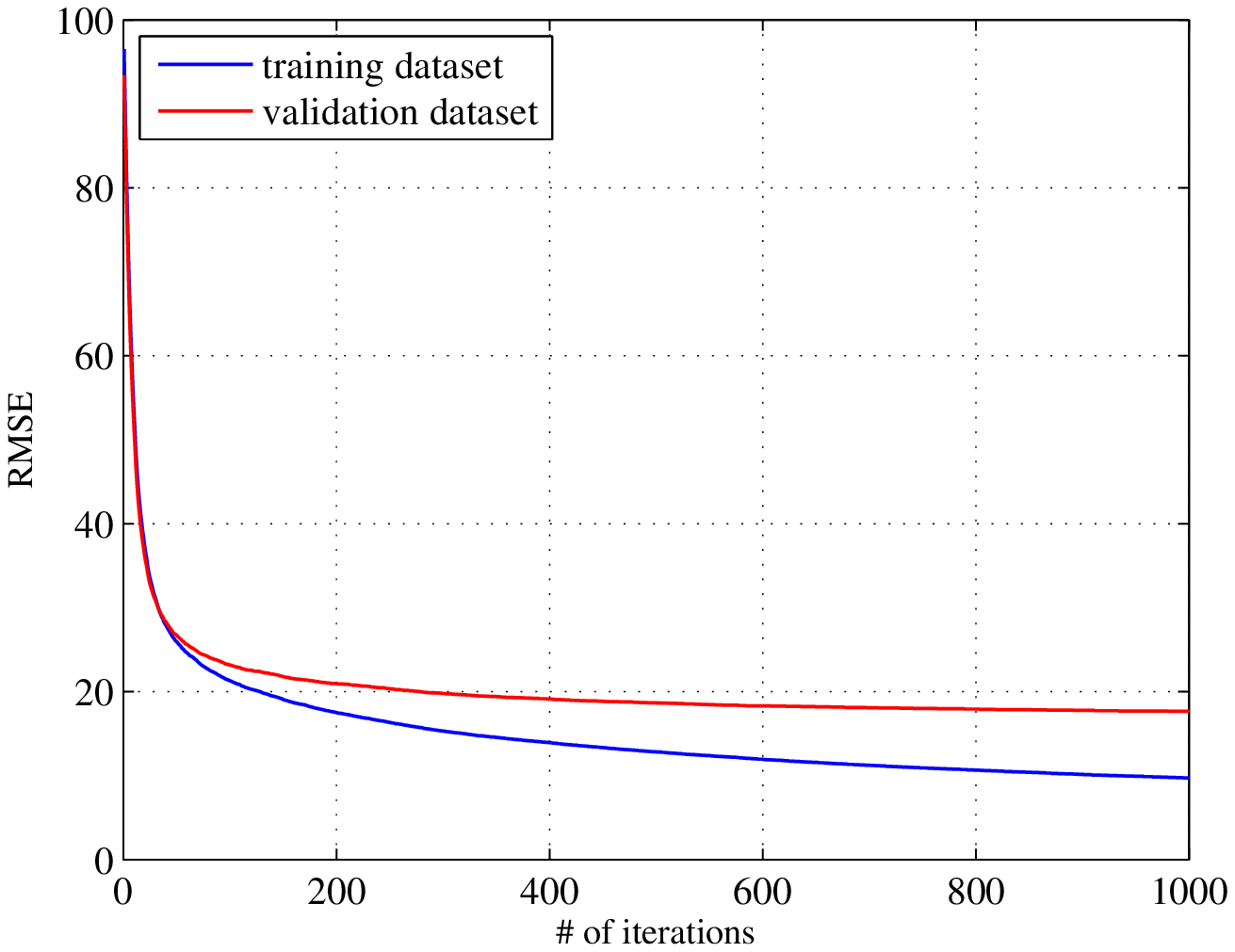}
		\caption{$\beta$=0.1, idepth=19, bag fraction=0.7, ntrees=1000}
		\label{fig:4e}
	\end{subfigure}
	\begin{subfigure}[b]{0.45\textwidth}
		\includegraphics[width=\textwidth]{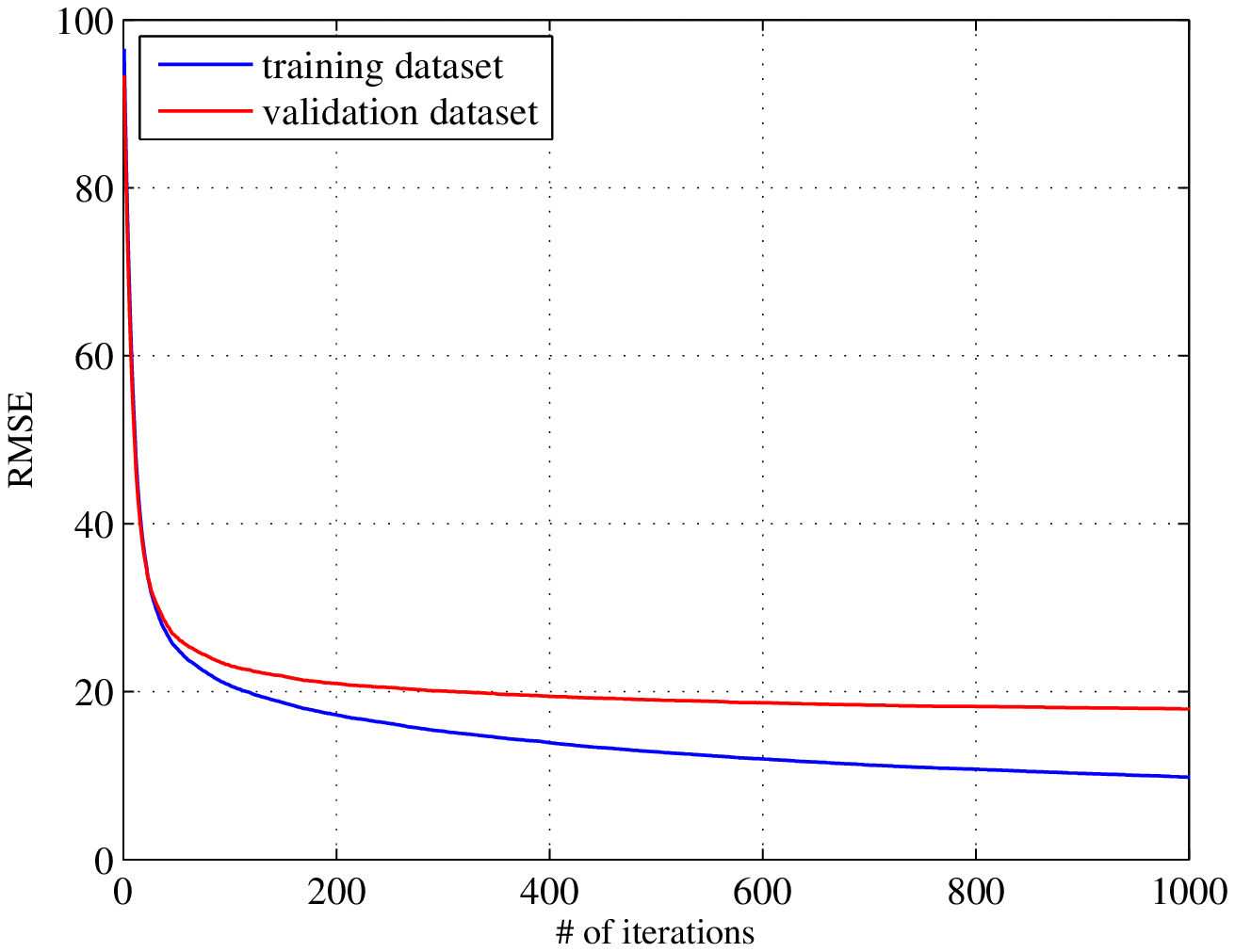}
		\caption{$\beta$=0.1, idepth=19, bag fraction=1.0, ntrees=1000}
		\label{fig:4f}
	\end{subfigure}
	\caption[]{Error curves for GBDT fitting on training and validation datasets for different parameter settings}
	\label{fig:4}
\end{figure}

In order to provide some general guidelines regarding the way a GBDT-based approach should be calibrated, we propose to apply the following steps. In the context of the estimation of the demand for ride-hailing and keeping in mind our large dataset, let us consider the default parameters with beta of 1, a bag fraction of 0.5 and an interaction depth of 1 and run the algorithm for 1000 iterations. To assess the performances of the model, the dataset is split into two independent training and validation sets according to the proportions 0.7 and 0.3 respectively. The best RMSE values with respect to the validation and training datasets are 56.05 and 58.07 respectively (\ref{fig:4a}). If we use the deep trees instead, we obtain 36.57 and 21.04 respectively(\ref{fig:4b}). This experiment shows us the influence of the complexity of the trees. Thus, deep trees are more interesting than stumps.

In some additional tests, we decreased the learning rate shrinkage to test the following values: 0.001, 0.01 and 0.1 by maintaining the same bag fraction and by using deep trees. Our tests have shown that for 2000 iterations, RMSE is equal to 16.68 with beta=0.1 instead of 26.12 with beta=0.001 while the number of iterations is superior (5000). For $\beta$=0.01, the RMSE is equal to 17.43 with even if the number of iterations reached 6000. As a result, a value of beta=0.1 provides the best performance in the context of our demand estimation for ride-hailing application.

With respect to the bag fraction, let us perform one test with the default value 0.5 (Figure \ref{fig:4d}) and another with 1.0 (Figure \ref{fig:4f}). The results reveal that, for a bag fraction of 1.0, the GBDT tends to over-fit the training dataset which mitigates its general predictive capabilities. RMSE with respect to the validation and training datasets are 17.94 and 9.82 respectively. By taking a middle value, one can reach a good trade-off between a good prediction accuracy and lower risk of over-fitting. In the case of a bag fraction value of 0.7 (Figure \ref{fig:4e}), the RMSE with respect to the validation and training datasets are 17.65 and 9.71 respectively. For a bag fraction of 0.5 (Figure \ref{fig:4d}), we obtain 18.00 and 10.24 respectively. As a result, a bag fraction of 0.7 provides the most satisfactory prediction accuracy.

\begin{figure}[H]
	\centering
	\includegraphics[width=\textwidth]{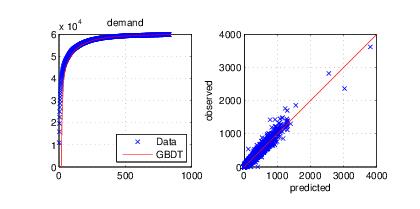}
	\caption{Comparison between the predicted and observed demand patterns based on GBDT}
	\label{fig:5}
\end{figure}

In conclusion, the best parameters can be defined as follows: $\beta$ (learning rate for shrinkage)=0.1, interaction depth using deep trees (also equal to the number of variables, i.e. 19), a bag fraction of 0.7 and an optimal number of iterations of around 5000 (when the error remains stable from an iteration to another). Note that the number of iterations can be increased once all the optimal parameters have been determined to achieve better predictions. For those settings, the GBDT presents an optimal RMSE of 16.41 with respect to the test dataset. In Figure \ref{fig:5}, the fit between the simulated and observed demand patterns clearly shows less spread compared to all the other mentioned approaches. The R-square value is 0.97 with a corresponding slope of 1.00. In addition, the points are uniformly distributed around the diagonal. Also, an important remark should be highlighted regarding the outliers. In fact, they correspond to times-lots where the demand is very important. In this regard, we observe that the GBDT approach succeeded in capturing and predicting all these specific demand time-slots as the three data points are close to the diagonal. 

Alternatively, the parameter tuning could have been set up via an iterative optimization problem for different parameter settings. However, in the context of this study, we do not recommend this approach. The dataset is too large, and given the important number of parameters, all the possible combinations can increase significantly and make the overall computational process relatively heavy. Instead, we recommended to have a look at the best practices adopted in the literature, e.g. bag fraction of 0.5 and a learning rate shrinkage of 0.1. In addition to the recommended values coming from the literature, one can tune the parameters by performing some tests and comparing the differences in terms of RMSE as presented. Such an approach is far better for sensitive analysis.

\subsection{Artificial Neural Network}
In the current study, a deep feedforward network with 2 hidden layers of 1,024 and 512 neurons respectively has been calibrated. Sigmoid activation functions and rectifier linear unit output function (clipping output to $max(0,y)$) is used in the model. In order to prevent the network from over-fitting, an inverted dropout with probability $p$ of 0.6 and 0.05 have been set up within the hidden layer inputs. The model has been trained for 1000 epochs. With respect to the input variables, they have been formatted into bit-wise binary units. At the end of the training phase, the dropout layer inputs are rescaled by $\frac{1}{(1-p)}$. Indeed, this practice allows the sparse layers to share statistical strengths learned during the training phase and no additional model trainings are necessary. Regarding the hyper-parameter tuning, we used a learning rate of 0.0001 with a learning decay of 1\% at each epoch and 0.9 momentum in our training on batch sizes of 150 samples. The lowest RMSE with respect to the validation dataset was obtained before the end of the training at epoch 473.

\begin{figure}[H]
	\centering
	\includegraphics[width=\textwidth]{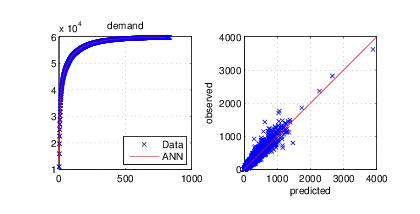}
	\caption{Comparison between the predicted and observed demand patterns based on feedforward ANN}
	\label{fig:13c}
\end{figure}

Figure \ref{fig:13c} presents the fit between the predicted and observed demand patterns based on ANN model. One can observe that ANN is capable of providing interesting predictive capabilities as high level demand patterns have been predicted correctly. However, for lower demand levels, i.e. around 1,000, minor spread can be observed. Globally, ANN has provided good statistics with a RMSE of 20.09, an R-square of 0.96 and a slope of 1.03.

\subsection{Detailed comparative study of the errors}

In addition to the aggregate comparison between the predicted and observed space-time demand, it is primordial to investigate the prediction accuracies for higher spatio-temporal resolutions. In this context, predicted DT, BDT, RF, ANN and GBDT-based demand patterns are classified across the 66 districts of the study area and systematically compared on the basis of the RMSE.

\begin{figure}[H]
	\centering
	\includegraphics[scale=1.]{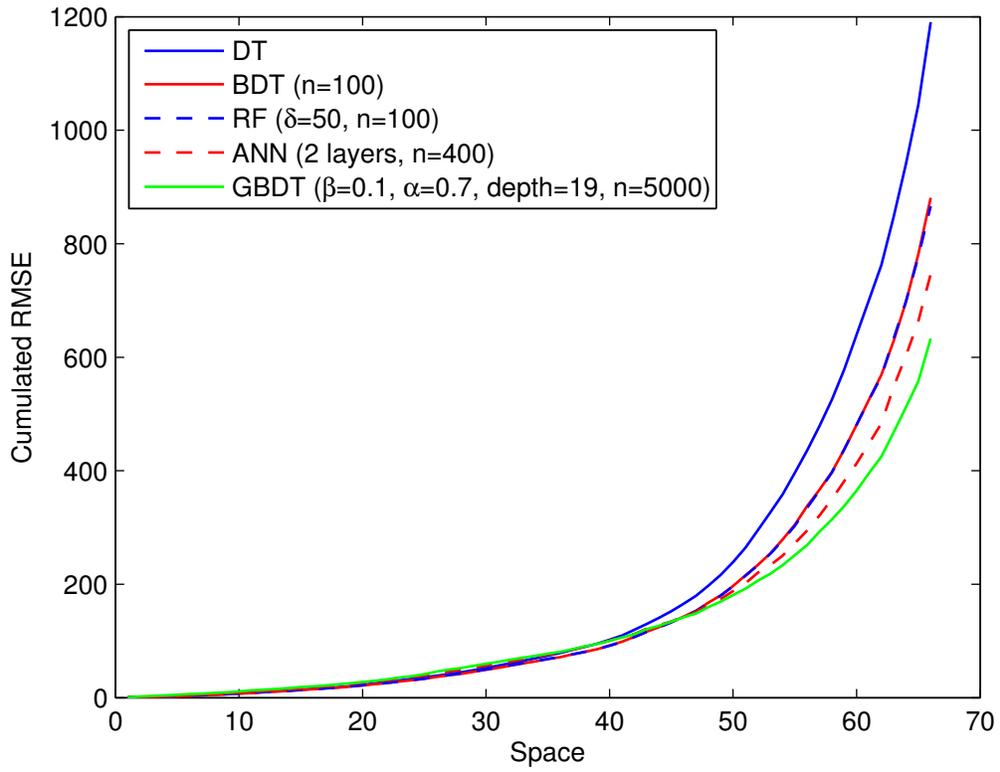}
	\caption{Comparison of the RMSE of DT, BDT, RF, ANN and GBDT based on space}
	\label{fig:13a}
\end{figure}

Figure \ref{fig:13a} shows that in the context of a more detailed spatial resolution, DT presents the highest cumulative RMSE while the evolution of the error related to BDT and RF are almost equivalent. ANN space-based error is ranked in-between RF and GBDT. GBDT shows once again the best prediction accuracies compared to the other methods as the spatial demand dynamics are characterized correctly. One could observe that the cumulative error curves are equivalent until the $40_{th}$ anonymized district, after which, the differences appear.

\begin{figure}[H]
	\includegraphics[scale=1.0]{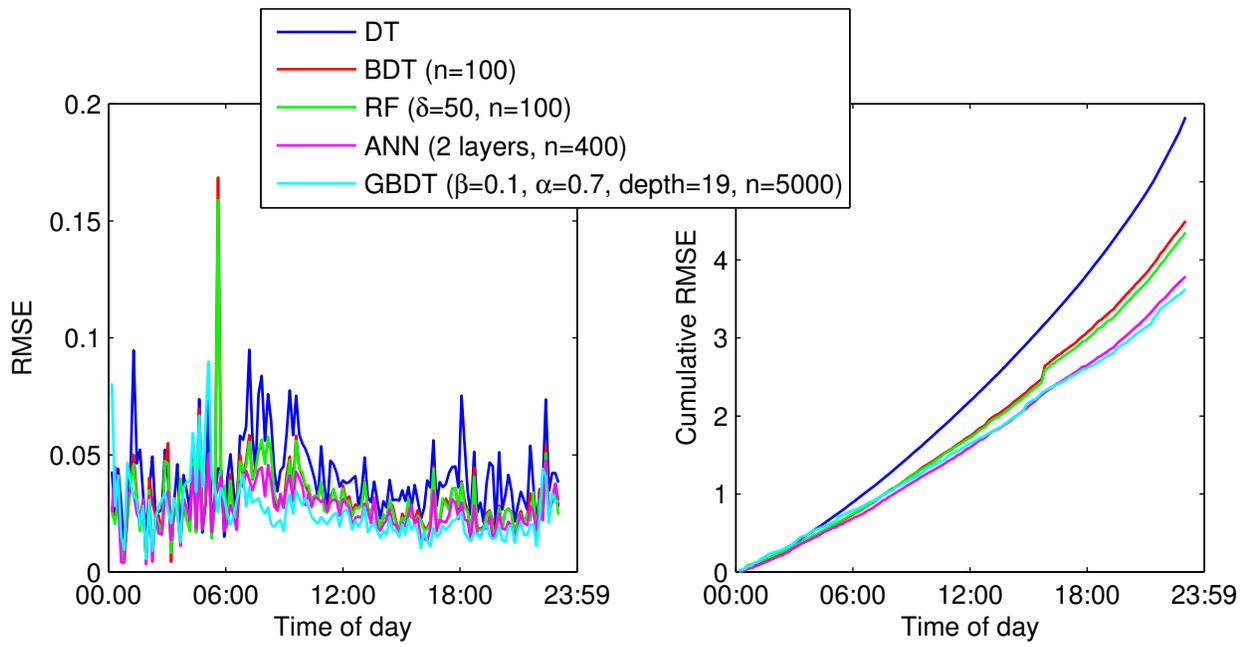}
	\caption{Comparison of the RMSE of DT, BDT, RF, ANN and GBDT based on time of day}
	\label{fig:13b}
\end{figure}

In Figure \ref{fig:13b}, the error curves are now based on time of day. GBDT presents the lowest RMSE along the day except for time period around 5:00am, ANN performs better. Therefore, ANN has the lowest cumulative error around 5:00am. As expected, DT has the worst RMSE. In the image of the trends observed in \ref{fig:13a}, BDT and RF are almost equivalent with a relatively lower RMSE. Indeed, the cumulative RMSE of RF can be distinguished under the one of BDT. Based on the cumulative error, GBDT also goes into the lead from 6.00pm compared to ANN.

\begin{figure}[H]
	\centering
	\includegraphics[scale=.9]{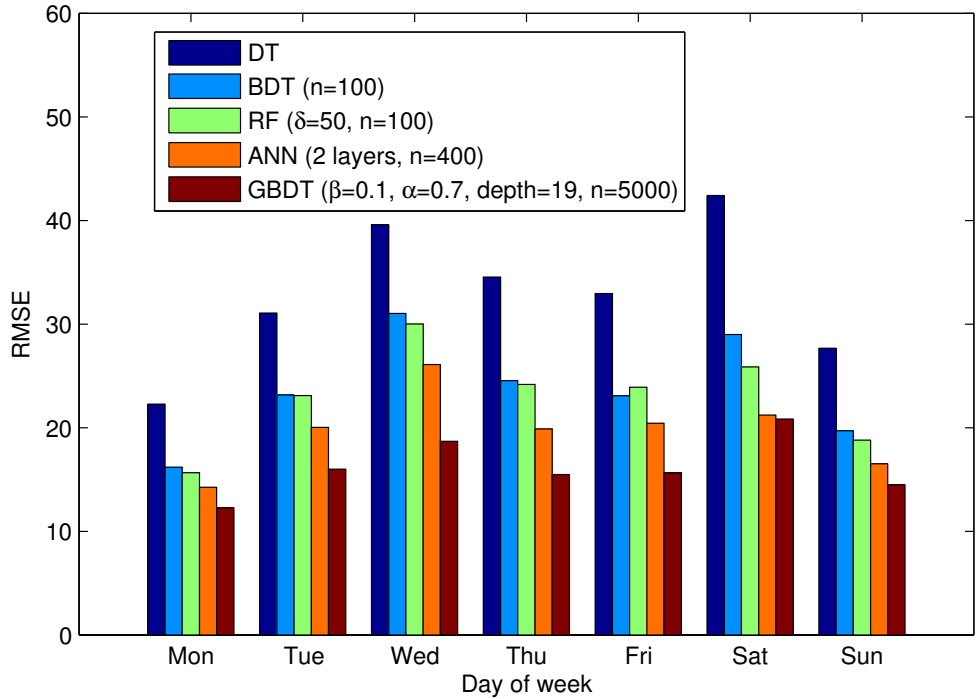}
	\caption{Comparison of the RMSE of DT, BDT, RF, ANN and GBDT based on day of week}
	\label{fig:13cc}
\end{figure}

In Figure \ref{fig:13cc}, we propose a temporal resolution based on days. DT presents once again very bad estimates with the highest RMSE values while BDT and RF-based approaches are almost equivalent for days 1, 2, 3, 4, 5 and 7. Regarding day 6, the deviation seems slightly more important. As expected, errors associated to ANN are systematically ranked at the $4^{th}$ position. GBDT algorithm outperforms undoubtedly all the methods for every single day of week.  


\section{Conclusions}

In this comparative study, we propose different nonparametric methods to characterize and forecast the short-term ride-hailing demand including spatio-temporal dynamics. Methodological features have been discussed in details as well as their relative performances by using various statistics. In addition to space-time, we have successfully managed to take into account the effects of other relevant factors associated to weather, pricing and traffic conditions. Relevant features are identified by running the RreliefF algorithm (an adapted version of the conventional relief algorithm for regression). The ensemble learning-based approaches show highest flexibility and efficiency in terms of data training. In addition, the computer run-time is very reasonable given the large dataset. In Section \ref{S:4} dedicated to data, we have highlighted the importance of the gap between demand and supply. The gap presents important disparities and depends on several factors especially the spatial location of the demand and, to a lesser extent, the time. In this context, the on-demand ride-hailing service providers need such type of efficient predictive models that demonstrate sufficient predictive capabilities for adjusting the imbalance between supply and demand.

Additionally, we have examined decision trees-based ensemble learning methods including a comparison with a two layers ANN approach. These two categories of nonparametric techniques for regression are the most popular. Regarding Support Vector Machine, after performing some tests, the results revealed that the algorithm was not adapted to handle the current ride-hailing problem. In addition, SVM for regression is computationally extremely expensive. Indeed, a run-time analysis revealed the existence of an exponential relationship between the increase of the number of observations and the run-time. At the same time, ensemble learning-based methods provide equivalent or even better predicted values while the computer run-time is relatively reasonable. Also, SVM includes two tuning parameters, i.e. resolution and kernel parameters, that need to be optimally determined based on a grid search technique. The optimization problem require several algorithm runs. In practice, the computer run-time would not be realistic. Thus, the SVM option had been quickly abandoned.

To the best of our knowledge, it is for the first time, the regression based machine learning models have been adapted and systematically tested on a real case study using the on-demand ride-hailing dataset of DiDi Chuxing (China). In this way, it is more convenient to assess the performance of the presented algorithms and choose the one that provide the best predictive capabilities for short-term demand forecasting. We envision that this study will provide a strong starting point for the use of machine learning based approaches as an alternative for the transportation demand forecasting models. 

In the current study, we preferred to opt for nonparametric models. Indeed, the nature and complexity of the variables and the size of the dataset cannot be truly encapsulated using parametric model fitting. For example, the spatial variable alone includes 66 categories. In addition, the demand estimation for ride-hailing is characterized by complex relationships between variables with a high level of dimensionality. Also, the shape of the ride-hailing demand is extremely left-skewed. Nonparametric techniques seem to be the most adapted to handle such type of problems. With respect to the modeling approaches, GBDT clearly provides the best performances in terms of general predictive capabilities.

The findings presented in this study are expected to have important implications on demand estimation for ride-hailing as on-demand service providers need to develop such tools capable of anticipating, within a short-term perspective, the spatio-temporal dynamics inherent to the demand. In doing so, the travelers requests can be fulfilled more efficiently, which would result in the mitigation of the gap.

In the context of further research, different aspects of the study can be improved. For example, the dataset describes the demand for the month of January 2016. For generalization purpose, the study can be extended to describe the demand including higher temporal scale, i.e. within a half year for example. In this regard, the size of the dataset could significantly increase, but we have highlighted the different procedures that can be adopted to mitigate the computer run-time using regularization, e.g. shrinkage, bagging, early stopping, random subspace. 

\section{Acknowledgements}

The research was funded by the ARC grant for Concerted Research Actions for project no. 13/17-01 entitled "Land-use change and future flood risk: influence of micro-scale spatial patterns (FloodLand)" and by the Special Fund for Research for project no. 5128 entitled "Assessment of sampling variability and aggregation error in transport models", both financed by the French Community of Belgium (Wallonia-Brussels Federation). This research was also supported by the FRQNT (Fonds de recherche du Qu\'ebec – Nature et technologies) for grant no. 202205 in the context of first author's research stay at Polytechnique Montr\'eal.

\section{References}
\bibliographystyle{model1-num-names}
\bibliography{sample.bib}

\newpage
\appendix

\section{Additional information regarding the fit of DT, BDT and RF with respect to the training dataset}

\begin{figure}[H]
	\centering
	\includegraphics[width=\textwidth]{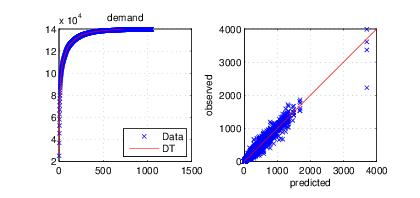}
	\caption{Comparison between the predicted and observed demand patterns based on DT}
	\label{fig:appA}
\end{figure}

\begin{figure}[H]
	\centering
	\includegraphics[width=\textwidth]{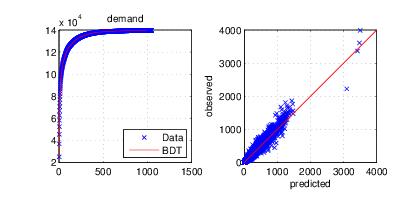}
	\caption{Comparison between the predicted and observed demand patterns based on BDT}
	\label{fig:appB}
\end{figure}

\begin{figure}[H]
	\centering
	\includegraphics[width=\textwidth]{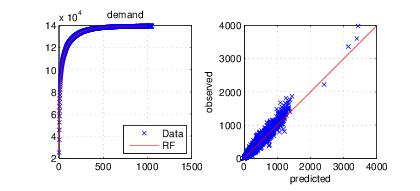}
	\caption{Comparison between the predicted and observed demand patterns based on RF}
	\label{fig:appC}
\end{figure}

\end{document}